\documentclass[a4paper,fleqn,authoryear,colorlinks=true,urlcolor=blue]{cas-dc}

\usepackage[authoryear]{natbib}
\usepackage{svg}
\usepackage{fixltx2e}
\usepackage{todonotes}
\usepackage{subcaption}
\usepackage{graphicx}
\usepackage{doi}
\usepackage{url}
\usepackage{etoolbox}
\usepackage{hyperref}
\usepackage{doi}
\usepackage{rotating}
\usepackage{float}
\usepackage{placeins}

\begin{document}
\let\WriteBookmarks\relax
\def\floatpagepagefraction{1}
\def\textpagefraction{.001}

\hypersetup{
    colorlinks=true,
    urlcolor=blue,
    citecolor=blue,
    linkcolor=blue
}

\shorttitle{Multi-Modal Machine Learning for Small Damage Detection}

\shortauthors{Khan et al.}

\title [mode=title]{Robust Anomaly Detection through Multi-Modal Autoencoder Fusion for Small Vehicle Damage Detection}

\author[1,3]{Sara Khan}\corref{cor1}
\ead{sakhan@uni-bremen.de}
\author[2]{Mehmed Yüksel}
\ead{mehmed.yueksel@dfki.de}
\author[1,2]{Frank Kirchner}
\ead{frank.kirchner@dfki.de}

\cortext[cor1]{Corresponding author}

\affiliation[1]{organization={Faculty of Mathematics and Computer Science, University of Bremen},
    city={Bremen},
    postcode={28359},
    state = {Bremen},
    country={Germany}}

\affiliation[2]{organization={Robotics Innovation Center, Deutsches Forschungszentrum für Künstliche Intelligenz},
    city={Bremen},
    postcode={28359},
    state = {Bremen},
    country={Germany}}

\affiliation[3]{organization={Engineering Software Communication, Robert Bosch GmbH},
    city={Renningen},
    postcode={71272},
    state = {Baden-Württemberg},
    country={Germany}}

\begin{abstract}
Wear and tear detection in fleet and shared vehicle systems is a critical challenge, particularly in rental and car-sharing services, where minor damage, such as dents, scratches, and underbody impacts, often goes unnoticed or is detected too late. Currently, manual inspection methods are the default approach, but are labour-intensive and prone to human error. In contrast, state-of-the-art image-based methods are less reliable when the vehicle is moving, and they cannot effectively capture underbody damage due to limited visual access and spatial coverage. This work introduces a novel multi-modal architecture based on anomaly detection to address these issues. Sensors such as Inertial Measurement Units (IMUs) and microphones are integrated into a compact device mounted on the vehicle's windshield. This approach supports real-time damage detection while avoiding the need for highly resource-intensive sensors. We developed multiple variants of multi-modal autoencoder-based architectures and evaluated them against unimodal and state-of-the-art methods. Our multi-modal ensemble model with pooling achieved the highest performance, with a Receiver Operating Characteristic-Area Under Curve (ROC-AUC) of 92\%, demonstrating its effectiveness in real-world applications. This approach can also be extended to other applications, such as improving automotive safety. It can integrate with airbag systems for efficient deployment and help autonomous vehicles by complementing other sensors in collision detection.
\end{abstract}


\maketitle
\section{Introduction}
\begin{figure}[ht]
    \centering
    \begin{subfigure}[b]{0.48\linewidth}  
        \centering
        \includegraphics[width=\linewidth]{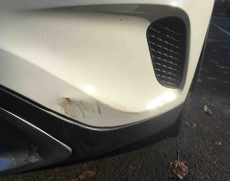}
        \caption{Cosmetic Damage}
        \label{fig:cosmetic}
    \end{subfigure}
    \hfill
    \begin{subfigure}[b]{0.48\linewidth}  
        \centering
        \includegraphics[width=\linewidth]{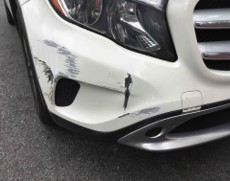}
        \caption{Significant Damage}
        \label{fig:significant}
    \end{subfigure}
     \caption{Examples of vehicle damages categorised by severity, adapted from \citep{Khan2023}.}
    \label{FIG:1}
\end{figure}

Automotive vehicles experience wear and tear during their use, requiring various strategies to detect and manage damage. These damages can be grouped into two main categories based on severity: significant damages, which are severe enough to affect a car’s operational functionality, and minor damages, which mainly alter appearance but leave functionality intact, as shown in Figure \ref{FIG:1}. Examples of minor damages include dents, scratches, and underbody wear \citep{Kurebwa2019ASO}.

In applications like ride sharing or car sharing, where multiple passengers use the same car, it is important to detect damage to ensure a smooth user experience and maintain accurate vehicle condition records. Reliable detection also enables accountability between users and service providers by helping determine when damage occurred and by whom, supporting transparent cost recovery and maintenance decisions. This work focuses primarily on minor damages, as they occur more frequently, can escalate into more significant issues if left untreated, and have a substantial impact on vehicle maintenance costs, resale value, and operational efficiency \citep{Lee2024AutomatedVD}. According to \citep{autorentalnews2021}, 12\% of renters feel that their vehicles were not in suitable appearance prior to renting, and 27\% feel that all damages were not properly recorded, as they are manually checked and written on a sheet.  

The current go-to approach remains manual inspection \citep{singh_automating_2019}, which is labour-intensive and prone to human error. To semi-automate the process, image-based methods have been deployed, allowing users to upload images of the car before and after rental. However, these methods are limited to static damage detection and restricted to visible damages \citep{Anderson2016}.

To address these limitations, we propose an alternative approach tailored for fleet and shared-vehicle scenarios. The method employs IMU sensors, traditionally used in airbag systems for high-impact detection \citep{shaikh2013air} and repurposes them to operate at lower thresholds, enabling the identification of minor impacts and subtle damage events.
In addition, microphones, widely used for impact and accident detection, are incorporated to provide complementary acoustic information. By combining these two sensors into a compact device mounted on a car’s windshield, we enable a more versatile solution. This setup, paired with data-driven models known as Small Damage Detection (SDD), facilitates damage identification even while the vehicle is in motion. While the primary focus is on shared-vehicle use cases, IMUs and microphones together enable low-latency data fusion, high sampling rates, and edge-computing capabilities for on-device processing \citep{Mumuni2021AdaptiveKF}. This method can also be extended to other applications. The SDD system shows great potential for enhancing automotive safety, particularly in collision scenarios. IMU sensors are already employed in airbag systems, where deployment is typically triggered by high-impact thresholds \citep{shaikh2013air}. These mechanisms can sometimes lead to unnecessary or mistimed deployments; for instance, at very high speeds, side airbags may deploy during a frontal collision, potentially harming occupants instead of protecting them. Although not the primary focus of this work, small-damage detection systems such as SDD could complement airbag decision logic by providing additional event-level information, thereby reducing false deployments and improving overall safety. Additionally, in autonomous vehicles, the SDD system can complement Light Detection and Ranging (LiDAR) sensors, which perform well at medium and long ranges but are less reliable at close range or in blind spots around the vehicle. This limitation is particularly relevant in low-speed manoeuvres, such as parking, where nearby obstacles or pedestrians may not be detected until it is too late. By incorporating IMU and audio sensing, the SDD system can bridge these ultra-close gaps, providing critical feedback on minor impacts or surface contacts that traditional sensors may miss \citep{chougule2023comprehensive}.

In this work, we present a machine learning-based approach utilising sensor hardware (comprising IMU sensors, a microphone, and a processor) mounted on a vehicle's windshield to detect small damages. We also address data-related challenges and the need for more expressive models, which motivated the development of our current autoencoder-based anomaly detection approach, as described in subsequent sections.

\begin{table*}[ht]
\centering
\caption{\textcolor{black}{Summary of recent research on vehicle damage detection using diverse sensor modalities.
This table highlights representative studies employing various sensors, such as audio, camera, IMU, and pressure to detect and classify vehicle damage or accidents across different application contexts.}}
\label{tbl1}
\begin{tabular}{ccc}
\toprule
\textbf{Citation} & \textbf{Sensor} & \textbf{Application} \\
\midrule
\citep{sammar2019} & Audio & Vehicle crash detection \\
\citep{scratch} & Camera images & Car scratch detection \\
\citep{Lee2024AutomatedVD} & Camera images & Car damage classification \\
\citep{dent} & Audio & Automotive dent detection \\
\citep{Dhieb2019AVD} & Camera images & Car damage detection and localization \\
\citep{harshani_image_2017} & IMU sensors & Accident and hazard detection \\
\citep{imu_accident} & Accelerometer, Pressure & Motorcycle accident detection \\
\citep{singh_automating_2019} & Camera images & Vehicle insurance claims \\
\citep{chiplunkar_deep_2021} & Camera images & Car damage classification and detection \\
\citep{choi} & Camera, Audio & Crash detection \\
\bottomrule
\end{tabular}
\end{table*}

\begin{figure*}[t] 
    \centering
    \includegraphics[width=\textwidth, keepaspectratio]{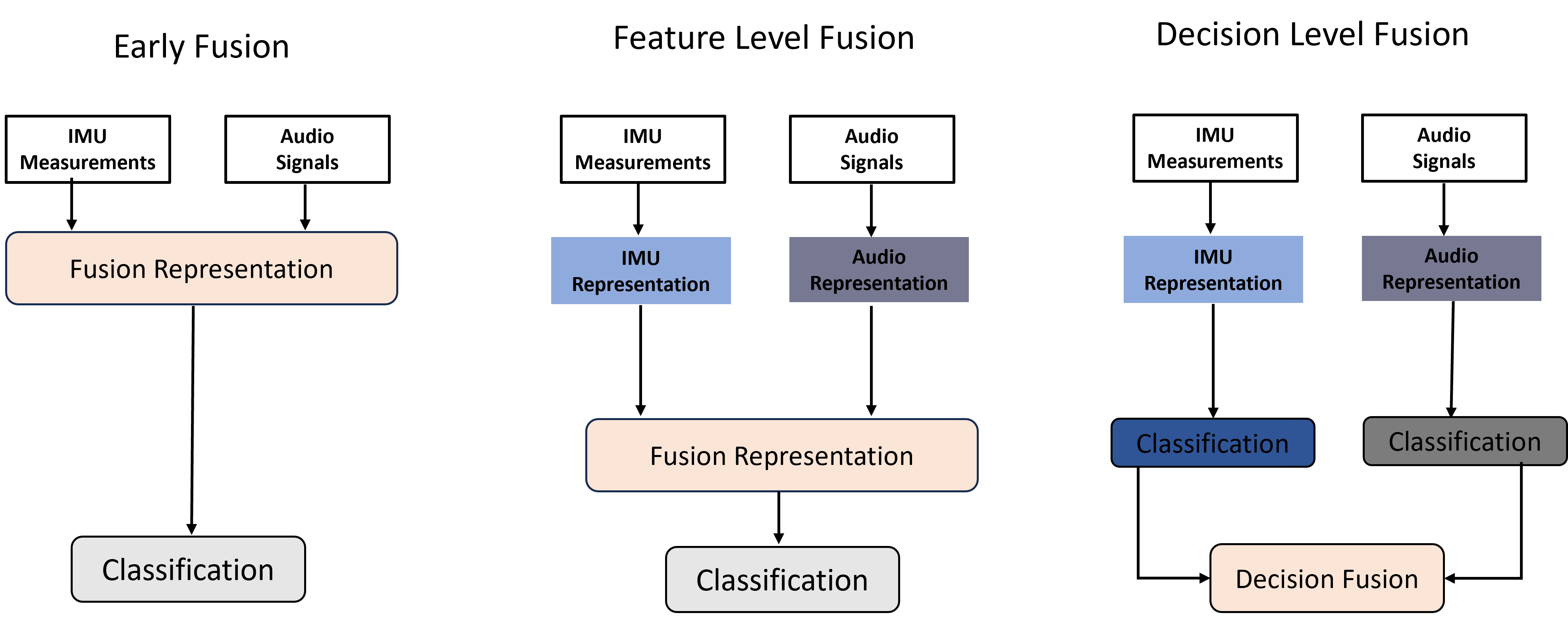} 
    \caption{\textcolor{black}{Schematic overview of multi-modal sensor fusion strategies. Early fusion combines raw IMU and audio signals before feature extraction; feature-level fusion merges learned features; and decision-level fusion integrates predictions at the output stage.}}
    \label{FIG:fusion}
\end{figure*}

\subsection{Related Work}\label{sec:related}
Damage detection has been extensively studied, particularly with a focus on camera-based sensors. Our earlier work \citep{Khan2023} provides a detailed exploration of these topics, including related areas like accident and impact detection. A summary is presented here.

\subsubsection{Camera-based Approaches}
Camera-based damage detection has been widely investigated using deep learning methods for accurate and efficient classification and localisation of damage. Techniques such as DenseNet-160 \citep{Lee2024AutomatedVD}, PANet \citep{singh_automating_2019}, MobileNet \citep{Waqas2020VehicleDC}, and Mask R-CNN \citep{masks_rcnn, masks_rcnn3} have been applied to car damage detection, while Darknet-53 with YOLO \citep{yolo, vanRuitenbeek2022CNNVehicleDamageDetection}, Inception-ResNetV2 \citep{Dhieb2019AVD}, and front-view damage detection using Convolutional Neural Network (CNNs) \citep{balci_front-view_2019} further demonstrate the versatility of deep learning. Innovative applications also include damage assessment and insurance evaluation \citep{VeHIDE2024}. Beyond deep learning, rule-based expert systems have been explored for severity and cost prediction \citep{harshani_image_2017}. However, challenges like high costs and limited scalability hinder real-time applications, especially in fleet and shared vehicle scenarios, underscoring the need for cost-effective solutions.

\subsubsection{Audio-based Approaches}
In contrast, audio-based approaches have shown promise in damage detection, particularly in accident scenarios. For instance, \citep{sammar2019} demonstrated the feasibility of differentiating crash sounds from other in-vehicle noises for damage detection. \citet{choi} integrated multi-modal data, including audio signals, for crash detection, achieving enhanced accuracy through a weighted ensemble of Gated Recurrent Unit (GRU) (\citep{gru}) and CNN models. \citet{dent} investigated the noise-to-signal ratio to detect dents in vehicles. These studies highlight the complementary potential of audio-based methods alongside camera-based approaches. Despite their potential, audio-based approaches have inherent limitations. Background noise, environmental variations, and differences in vehicle acoustics pose significant challenges in reliably distinguishing damage-related sounds from other noises in vehicles. Consequently, integrating audio data with complementary sensing modalities, such as vision or IMU measurements, can detect even subtle vibrations and accelerations that might not produce audible noise but still indicate an impact.

\subsubsection{IMU-based Approaches}
IMU sensors have also been explored for damage detection. \citet{harshani_image_2017, accidentodo} demonstrated the use of IMU sensors for detecting accidents and hazards in two- and four-wheeler vehicles. Studies \citep{motorcycle, motorcycle2} employed acceleration sensors for crash detection in motorcycles, while other works \citep{imu_accident, accident2} focused on accident detection more broadly.

\citet{msle} applied adaptive Kalman filtering for crash detection using multi-sensor decision fusion. \citet{imu_accident} developed an application that combined an external pressure sensor with GPS and accelerometer data for accident detection. \citet{gont} emphasised the potential of piezoelectric sensors for real-time detection of minor damages, motivating the development of data-driven systems.

Table~\ref{tbl1} summarises recent studies on damage detection using different sensor configurations. We observed that camera-based methods have dominated research so far; however, due to their limitations, they are not well suited to our problem, opening avenues for alternative approaches based on audio or IMU data.

\subsection{Open Challenges}
\begin{itemize}
    \item Camera-based methods are less effective in detecting underbody damage or when vehicles are non-stationary, due to occlusions, spatial dependence, and high computational load.
    \item Audio-based methods struggle with background noise interference and the difficulty of distinguishing subtle impact sounds.
    \item IMU-based approaches can detect impacts but may not provide sufficient context to differentiate minor damages from normal vehicle movements.
    \item There is a lack of multi-modal fusion strategies that effectively combine audio and IMU data to enhance detection accuracy.
\end{itemize}

\subsection{Novel Contributions}
This work addresses the above challenges by proposing a data-driven, multi-modal approach for real-time minor-damage detection using IMU and microphone sensors.

\begin{itemize}
    \item A mid-fusion autoencoder architecture (MAA3) with pooling improves multi-modal integration, achieving a ROC–AUC of 0.92.
    \item Evaluation of real-world, windshield-mounted IMU and audio sensor data demonstrates that non-visual sensing can support damage detection without reliance on cameras.
    \item Mono-modal audio performs comparably to fused models for certain damage types, highlighting its standalone potential.
    \item A semi-supervised anomaly detection formulation reduces the labelling effort required during model training.
\end{itemize}

\section{Background} \label{sec:background}
This section covers two foundational concepts used in this work: multi-modal sensor fusion and autoencoders. Although both are well-established in the broader machine learning literature, their reintroduction here is essential to contextualise their specific roles within the proposed framework. Multi-modal fusion is leveraged to integrate IMU and audio data, addressing the limitations of single-sensor approaches in detecting minor damage. Autoencoders, on the other hand, are conventionally used for image reconstruction or compression tasks but are adapted here for anomaly detection to identify subtle damage events. Their unsupervised feature-learning capability makes them particularly suitable for scenarios where labelled data are scarce or unavailable.

\subsection{Sensor Modalities and Fusion Strategies}  \label{sec:3.1}
This section discusses different types of strategies using single or multiple modalities. We classify them into two categories: mono-modality and multi-modality sensor fusion approaches. Mono-modality refers to the use of a single type of sensor to perceive and interpret an environment. Multi-modality refers to the integration of multiple sensor modalities to capture diverse aspects of a phenomenon or environment. Each sensor modality offers unique strengths. For instance, cameras provide rich visual information about the environment, and IMUs capture vehicle motion and orientation. Audio sensors can detect sound signals, enabling applications such as collision detection and anomaly identification. 

Despite their advantages, mono-modal systems face challenges in robustness and reliability due to the constraints of a single sensor. In multi-modal fusion (\citep{Yeong2021}), data is combined from different sensors such as cameras, radar, IMUs, and audio sensors, allowing vehicles to obtain comprehensive and robust information about their surroundings. \\
\textcolor{black}{As illustrated in Figure \ref{FIG:fusion}, three primary fusion strategies are typically used to integrate data from multiple modalities \citep{doe2023multi}}:\\
\textbf{Early Fusion}: Combine raw data from multiple sensors before processing. This method leverages complementary sensor strengths but can be computationally demanding due to the need for data alignment and preprocessing. \\
\textbf{Feature-Level Fusion}: Integrates extracted features from each sensor and computes their individual feature representation, and is then fused into a common representation before feeding into the model. This approach balances complexity and efficiency, enhancing classification while reducing computational load compared to early fusion. \\
\textbf{Decision-Level Fusion}: Merges decisions or predictions from individual sensors after independent processing. Techniques like voting or averaging improve overall accuracy but may not fully utilise the synergy between sensors.\\
\textbf{Hybrid Fusion}: Combine elements of early, feature-level, and decision-level fusion to optimise accuracy and robustness. Although it offers comprehensive benefits, it adds complexity to system design.

\subsection{Autoencoders as Automated Feature Extractors}
Feature extractors play an important role in machine learning, as the quality of extracted features directly affects model performance. \textcolor{black}{Early approaches relied on handcrafted features, which often introduced bias and limited generalisation. Classical models, such as Principal Component Analysis (PCA) \citep{pca_linear}, automated this process to some extent, but they captured only linear dependencies in data. Convolutional layers in supervised networks, such as CNNs, are capable of modelling more complex, non-linear patterns; however, they typically require large labelled datasets. Non-linear autoencoders bridge these approaches by extending PCA to non-linear representations and, when needed, incorporating convolutional layers.} Importantly, they can be trained in an unsupervised manner, enabling the learning of compact and expressive feature representations from raw IMU and audio signals without reliance on extensive annotation.

Autoencoders \citep{hinton2006ae} are a class of neural network models used for unsupervised learning and dimensionality-reduction tasks. The fundamental principle behind autoencoders is to learn a compressed representation, or encoding, of the input data, typically with lower dimensionality than the original input. This encoding is then decoded to reconstruct the input data as accurately as possible.

\begin{figure}
    \centering
    \includegraphics[scale=0.34]{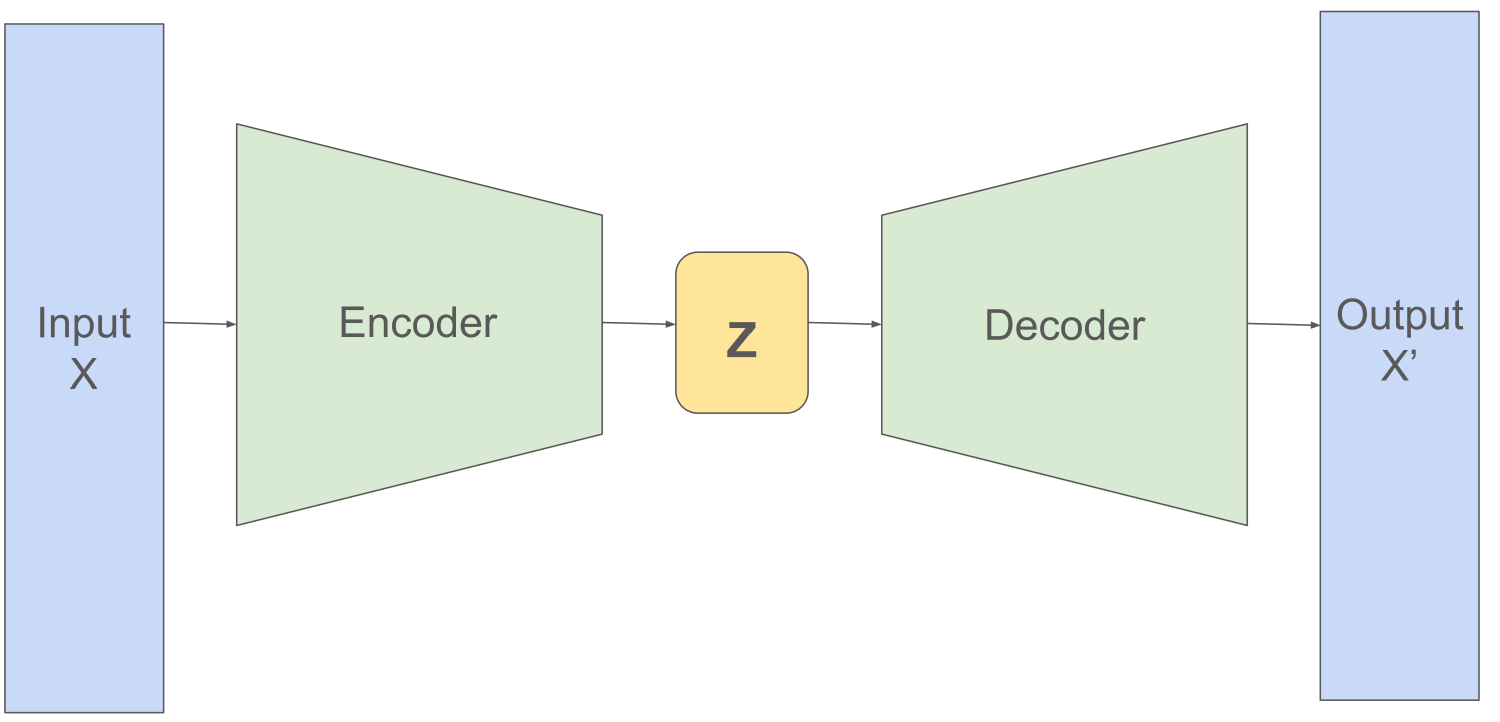}
    \caption{Structure of an autoencoder used for anomaly detection. \textcolor{black}{The encoder compresses input signals (X) into a latent representation (Z), while the decoder reconstructs them ($X'$), enabling damage-related anomaly detection through reconstruction error.}}
    \label{FIG:ae}
\end{figure}

Autoencoders consist of two main components: an encoder and a decoder. The encoder maps the input data to a latent-space representation, capturing essential features and patterns. \textcolor{black}{The decoder then reconstructs the original input from this encoded representation, as shown in Figure \ref{FIG:ae}, where the green blocks represent the encoder and decoder, and the yellow block denotes the latent space (Z). During training, autoencoders learn to minimise the reconstruction error, thereby producing a compact and informative latent representation} \citep{hinton2006ae}.

Several variations of autoencoders exist to address different learning challenges \citep{autoencodervar}. Sparse autoencoders impose a sparsity constraint on the hidden layers, encouraging the network to activate only a small subset of neurons, which results in a more efficient and compact representation. Denoising autoencoders are trained to reconstruct clean input data from noisy versions, enhancing their robustness and feature-extraction ability. Variational autoencoders (VAEs) \citep{vae} extend the basic framework by introducing a probabilistic formulation, allowing for the generation of new data by sampling from learned distributions in the latent space. Convolutional autoencoders \citep{conv_ae} apply convolutional layers, making them particularly suitable for time-series or image-processing tasks such as compression and denoising. Lastly, contractive autoencoders \citep{contr_ae} add regularisation to the latent space, promoting stability in the learned representation and ensuring that small input perturbations do not drastically affect the output. Each of these autoencoder types offers unique advantages depending on the specific task and the nature of the data being processed.

\section{Methodology} \label{sec:algorithms}
This section provides an overview of various methodologies, ranging from fully manual techniques to fully automated systems. It begins by exploring manual inspection methods and then transitions to data-driven approaches.

\subsection{Manual Inspection Methods}
Manual inspection methods for vehicle damage detection involve a human inspector visually inspecting the vehicle to identify and assess any damage. This process typically includes several steps: pre-inspection preparation, visual inspection, documentation, damage assessment, and report generation \citep{cardamage2020}. Manual inspection methods have long been the conventional approach to assess damage in car rental applications. For example, when renting a car from car rental companies, customers are typically required to manually inspect and document all minor or cosmetic damage before renting the vehicle. Upon returning the vehicle, this documentation is used to reassess any additional damage, and customers are charged accordingly for any new damage. In addition, the subjective nature of manual inspections often leads to disputes between different stakeholders, such as vehicle owners, repair shops, and insurance companies. The study \citep{singh_automating_2019} also shows that manual inspections can be less efficient, requiring more time and resources compared with automated methods. A single inspection may cost between 30 and 200 dollars and may take between 3 and 5 hours \citep{inspektlabs2024}. These limitations underscore the need for more objective, consistent, and efficient damage detection methods.

\begin{figure}[h]
    \centering
    \includegraphics[width=\columnwidth, keepaspectratio]{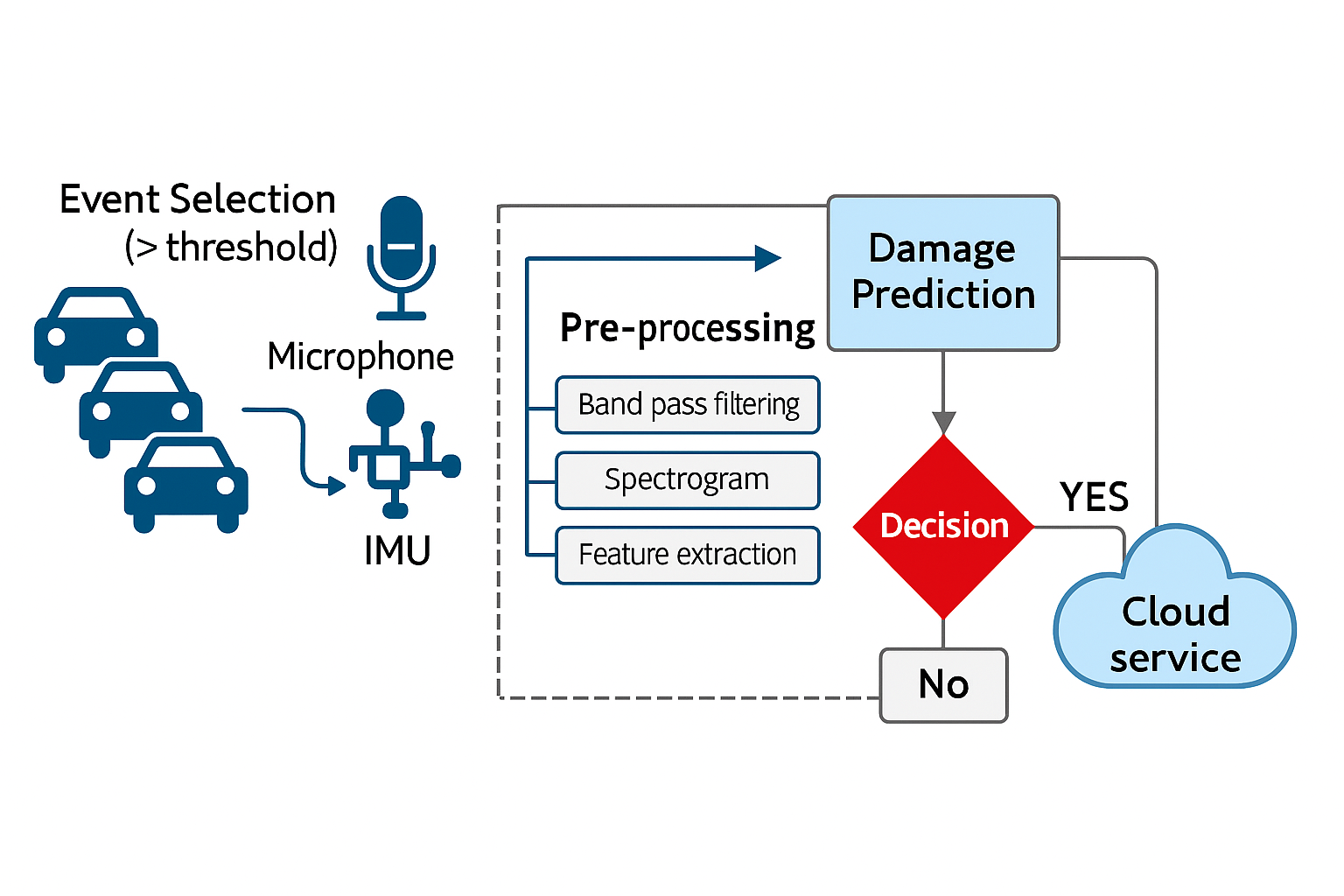}
    \caption{Architecture of the Small Damage Detection (SDD) pipeline, showing sensor input, preprocessing, model inference, and cloud-based logging for confirmed events. The system triggers data capture when acceleration thresholds are exceeded.}
    \label{fig:sdd_framework}
\end{figure}

\subsection{Data-driven Methods}
To address the limitations of manual inspection methods, this section focuses on data-driven approaches. We describe SDD, data collection and setup, data challenges, and a machine learning based approach. The modelling process follows the CRISP-DM \citep{crisp} framework.

\subsubsection{Small Damage Detection (SDD)}
SDD involves a mechanism for identifying potential damage using sensory data. This mechanism operates on an event-driven strategy, initiated during vehicle rides. These events are broadly categorised into two groups: events corresponding to damages and those representing non-damage scenarios. \textcolor{black}{As illustrated in Figure \ref{fig:sdd_framework}, the process begins with IMU and microphone inputs that trigger event selection when acceleration thresholds are exceeded. The data are then pre-processed through band-pass filtering, spectrogram generation, and feature extraction before being passed to the damage prediction model.} When sensory data exceed a predefined threshold, the system captures relevant information within a designated time window. The raw data undergo a pre-processing phase and are subsequently utilised by a machine-learning model that has been trained to perform classification. This model predicts whether the given scenario indicates damage or non-damage (background). In instances of detected damage, relevant information is transmitted to a cloud service to facilitate further action.

\begin{figure*}[ht]  
    \centering
    
    \begin{minipage}[t]{0.49\textwidth}
        \centering
        \includegraphics[width=\textwidth]{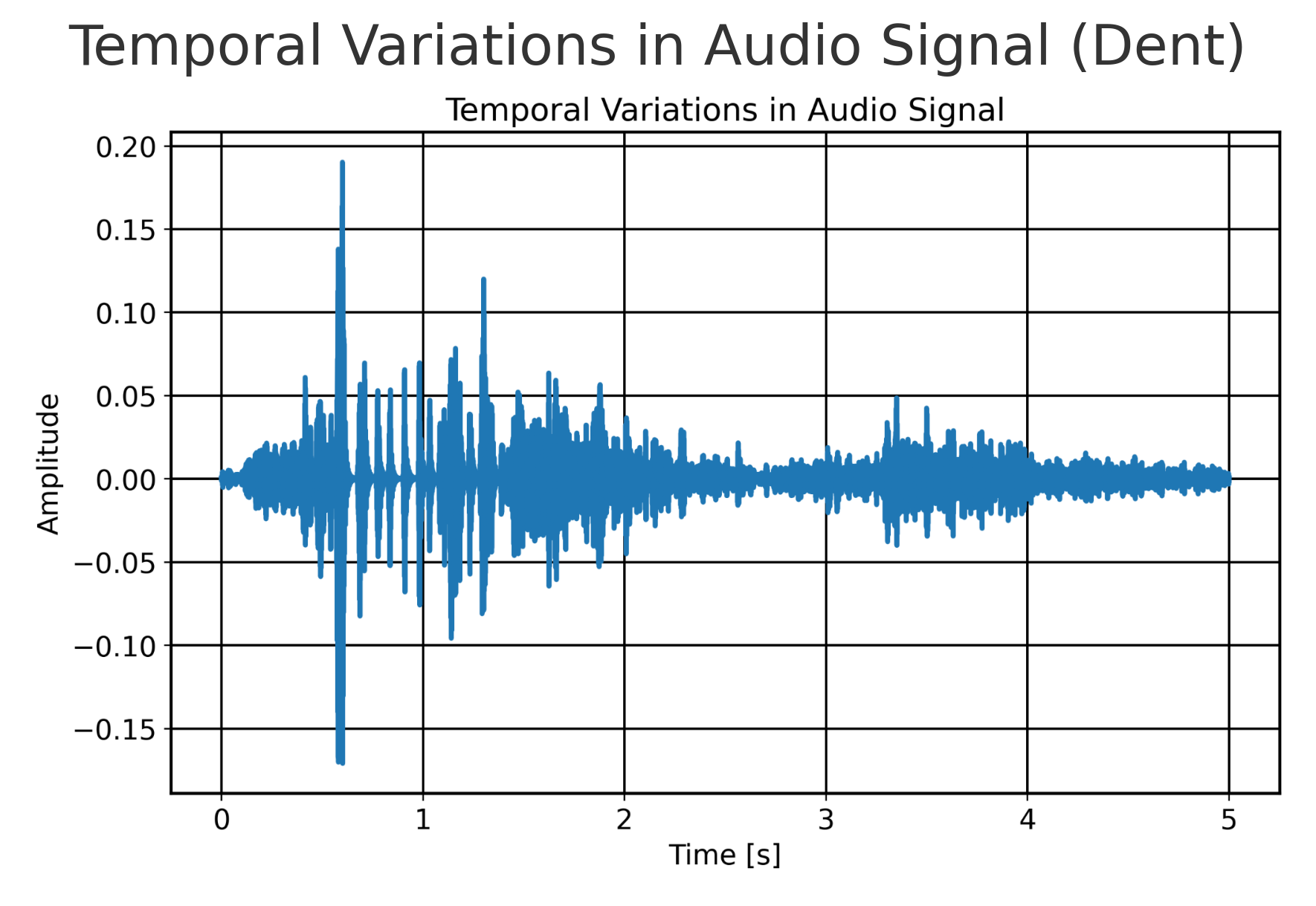}
    \end{minipage}
    \hfill
    \begin{minipage}[t]{0.49\textwidth}
        \centering
        \includegraphics[width=\textwidth]{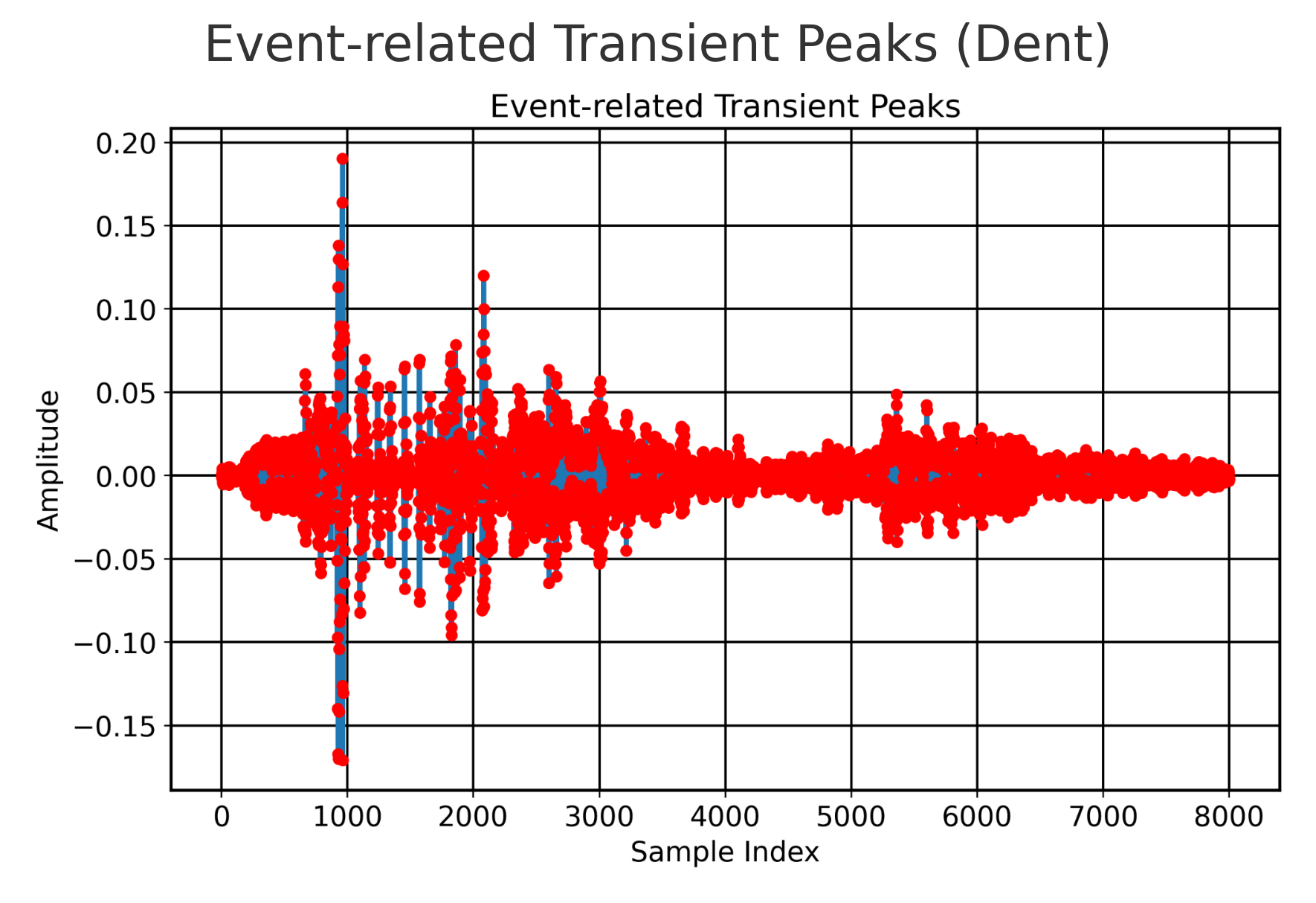}
    \end{minipage}
    
    \vspace{6pt}  
    
    \begin{minipage}[t]{0.49\textwidth}
        \centering
        \includegraphics[width=\textwidth]{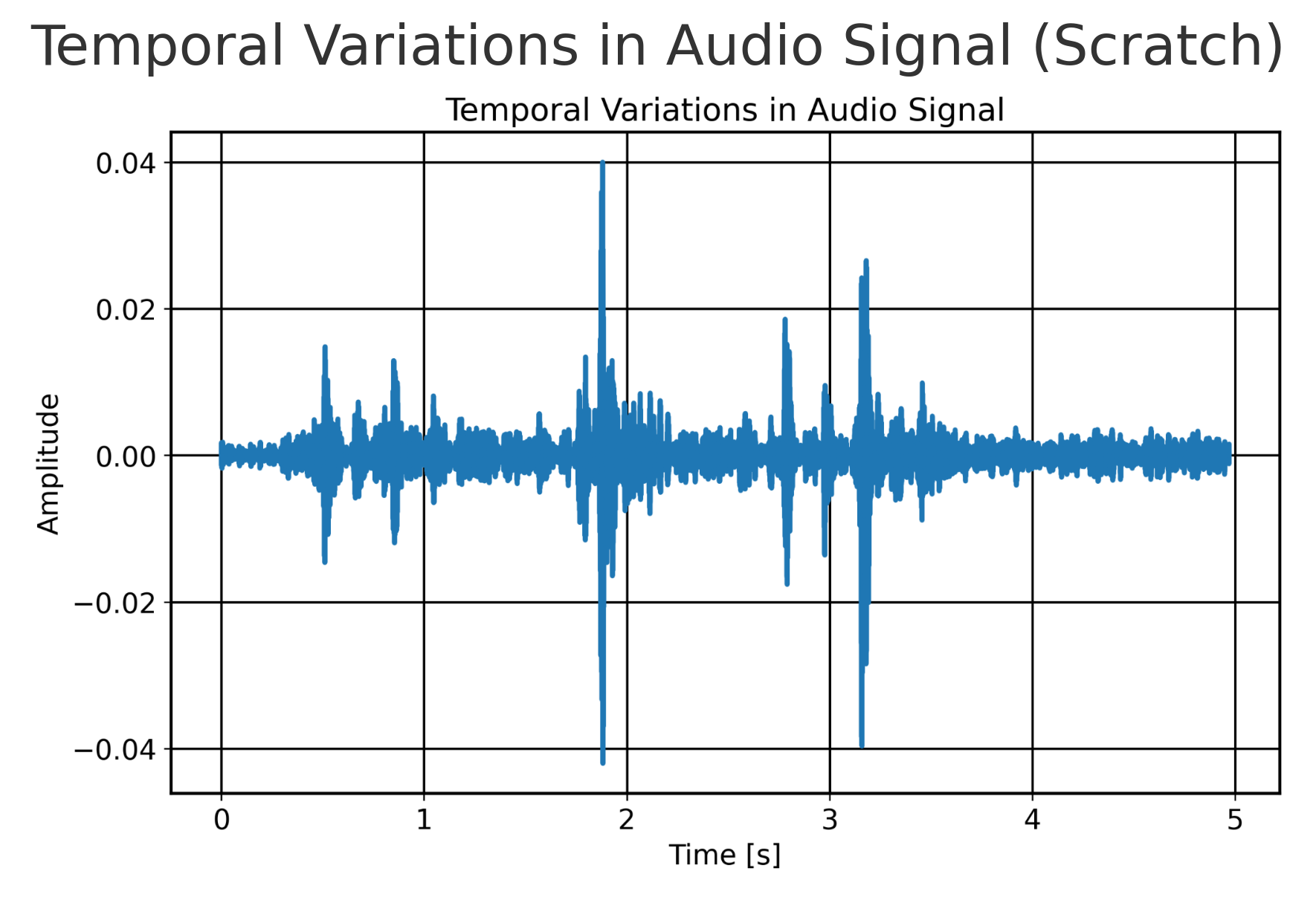}
    \end{minipage}
    \hfill
    \begin{minipage}[t]{0.49\textwidth}
        \centering
        \includegraphics[width=\textwidth]{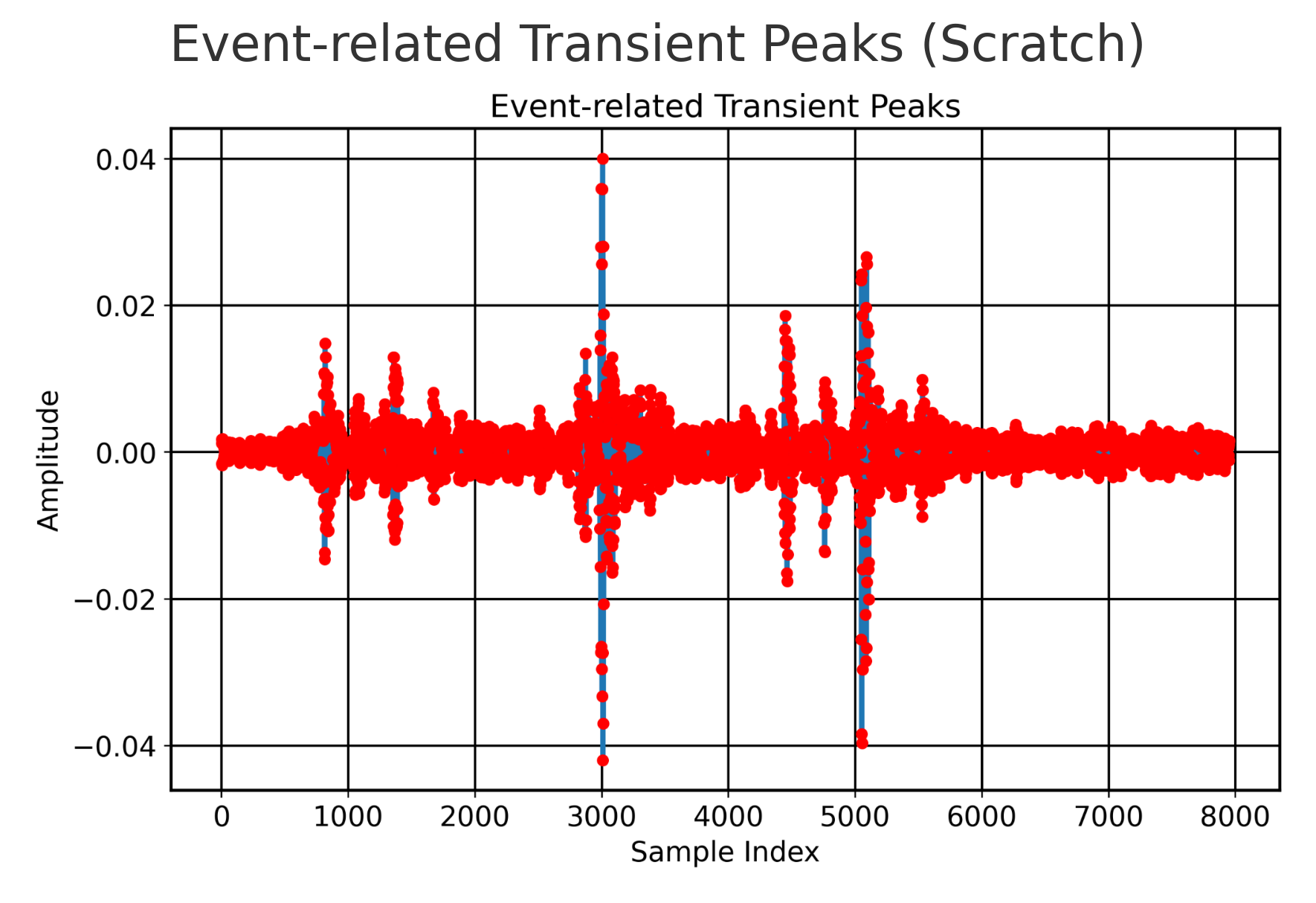}
    \end{minipage}
  
    \caption{\textcolor{black}{Comparative feature characteristics of two damage types: (a) dent (top) and (b) scratch (bottom). The left column shows temporal variations, and the right column shows event-related features derived from accelerometer and audio signals.}}
    \label{fig:combined}
\end{figure*}

\subsubsection{Dataset}
The dataset was collected through a structured data collection process, which involves conducting experiments for various damage and non-damage events. For example, if a vehicle hits another vehicle, it is classified as a damage event, whereas if a door closes, it is classified as a non-damage event. The experimental setup includes mounting the hardware on the car's windshield \citep{Bosch2024}, using a dashcam to record the experiments, assigning a driver to perform the experiments, and having a co-driver label the data using a graphical user interface application. 
The dataset is derived from 500 experiments, each comprising a complete set of the above-mentioned steps involving specific damage and non-damage scenarios. Raw sensory data from these experiments underwent pre-processing and filtering. The dataset consists of data from IMU and microphone sensors, resampled at 8000 Hz and 1600 Hz, respectively. To further process the data, acceleration data is subjected to a low-pass filter with a cutoff of 218 Hz, while audio data undergoes three band-pass filters of different ranges before feature computation.
\begin{figure}[h] 
  \centering %
  \includegraphics[width=0.5\textwidth]{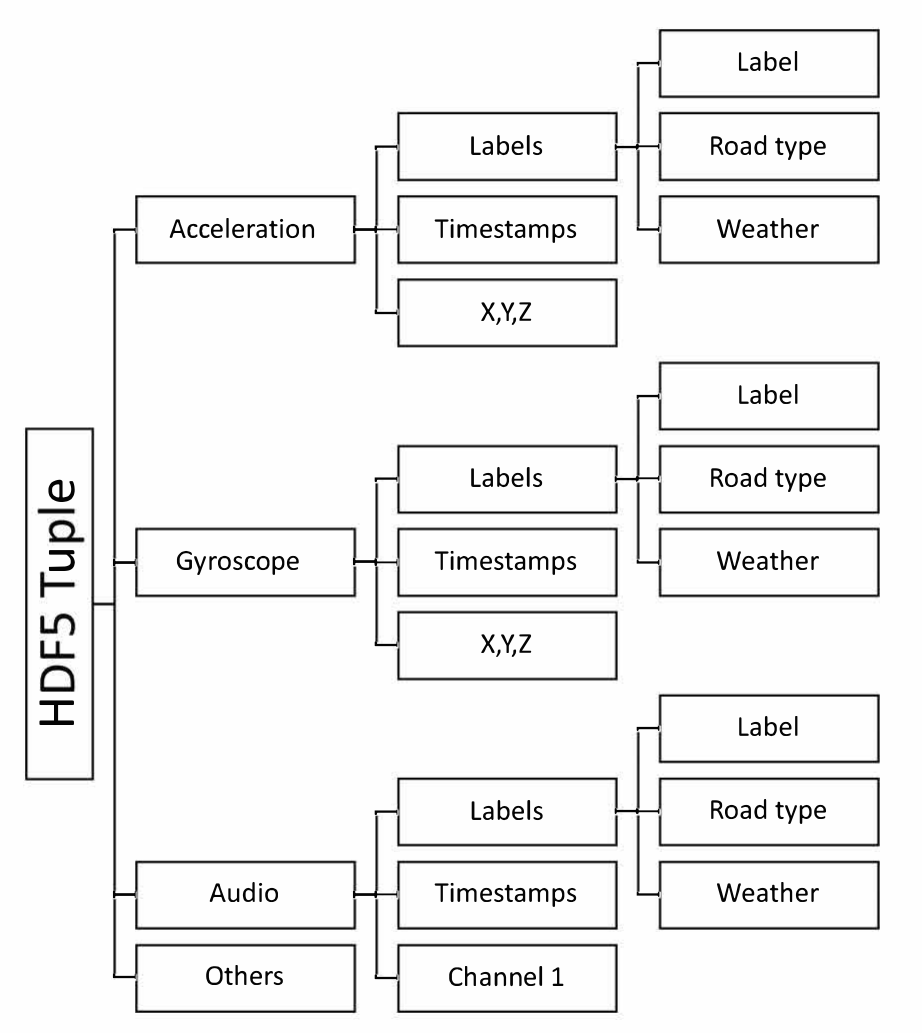} 
  \caption{Hierarchical storage of an event sample in HDF5 format. The structure includes accelerometer, gyroscope, and audio channels, along with contextual metadata. In this work, only the accelerometer and audio streams are used for modelling.}
  \label{fig:datalabel}
\end{figure}
Data collection spanned varied weather conditions, including sunny, cloudy, and rainy days, and diverse road surfaces such as asphalt, stone roads, gravel, mud, dirt, and snow, particularly for acquiring data from moving vehicles. This comprehensive coverage aims to increase the model’s robustness and adaptability. Multiple drivers operated different vehicle brands and models during the recordings, including Audi A5 (stationary and moving), BMW i3, BMW X1, BMW 5 series, Mercedes A45 (stationary and moving), Mercedes GLA, Mini, Smart, Volkswagen Polo, and Volkswagen Tiguan, predominantly in stationary conditions. 
The dataset presents several challenges. Firstly, similar sensor outputs from different events, such as minor damages and regular driving behaviours, can appear nearly identical, making differentiation difficult. For example, acceleration patterns from a non-damage event like normal driving can closely resemble those from a damage event, such as a scratch. Secondly, events that do not cause damage, such as a door closing, can produce signals that misleadingly suggest a dent or scratch, leading to false positives. As a result, a simple rule-based strategy is not suitable for this problem. Lastly, the dataset is heavily skewed towards non-damage instances, with a ratio of 40:1, causing models to favour predictions of no damage and overlook actual damages. Examples of challenges 1 and 2 are discussed in \citep{sara2024}. These challenges emphasise the need for a machine learning model capable of generalising to novel scenarios.

Upon investigating the audio data collected during in-field test sessions, we gained valuable insights into the acoustic environment surrounding the vehicle in various driving scenarios.

Figure \ref{fig:combined} presents a comparative analysis of two types of damage events: dents (top row) and scratches (bottom row). The left column illustrates the temporal variations of the corresponding audio signals (in blue), while the right column depicts the event-related transient peaks (in red) derived from audio features. \\
\textbf{Temporal Variations}: The temporal waveform analysis reveals significant differences between the dent and scratch events. The dent signal demonstrates a sharp, transient peak in amplitude, followed by a rapid decay, characteristic of a sudden impact. Conversely, the scratch signal presents a more prolonged and irregular waveform, indicative of a sustained interaction between surfaces. The amplitude fluctuations are less abrupt, corresponding to the continuous scraping noise. \\
\textbf{Event-related Features}: The event-related plots further distinguish the dent from the scratch. The dent event is marked by a distinct and isolated transient peak, which stands out against the background noise, highlighting its sudden and intense nature. On the other hand, the scratch event shows multiple peaks and valleys, indicating a more complex and sustained interaction. These features are critical in differentiating between the two damage types, as they provide clear visual cues about the nature of the events.  \\
Overall, Figure \ref{fig:combined} underscores the distinct temporal and event-related characteristics of dent and scratch events. The distinct signal characteristics serve as robust indicators for damage detection, and the careful selection of relevant acoustic features is vital for the effective application of machine learning algorithms in this context. To support such feature-based analysis, Figure \ref{fig:datalabel} illustrates the underlying data organisation used in this work, showing how each experiment is stored with synchronised IMU and audio streams, along with associated metadata such as timestamps, sensor axes, and contextual details like weather, road type, and damage category.

\subsection{Methodologies based on fusion strategies}
This section describes the different multi-modal frameworks developed in this study. Two modalities were used: acceleration and audio. Their integration was examined to evaluate the effect on performance. 
\subsubsection{Mono-modal approaches}
\textcolor{black}{The initial set of approaches is mono-modal, wherein models are constructed solely based on either acceleration or audio data. These serve as baseline models for comparison with multi-modal architectures.}

\begin{figure*}[htbp]
    \centering
    \includegraphics[width=0.9\textwidth]{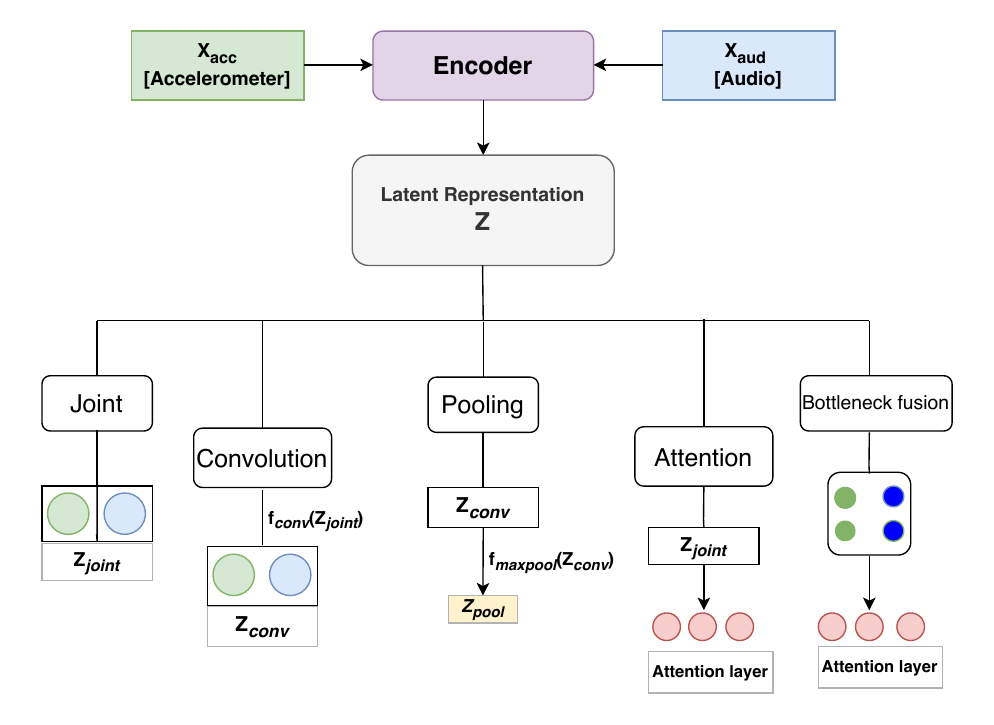}  
    \caption{Different variants of multi-modal architectures using feature-level fusion. (a) Joint Variant, (b) Convolutional Variant, (c) Pooling Variant, (d) Attention Variant, (e) Bottleneck Variant}
    \label{fig:var}
\end{figure*}

\subsubsection{Multi-modal Approaches}
To further improve performance, we develop multiple multi-modal architectures that integrate both acceleration and audio modalities. Various fusion strategies, early, mid, and late fusion, were explored in Section \ref{sec:3.1}. In this study, we focus on mid-fusion, which allows effective cross-modal interaction while maintaining modality-specific features. It balances the integration of complementary signals from multiple modalities with manageable computational complexity.

The input data, denoted as \( X_{\text{acc}} \) and \( X_{\text{audio}} \), are processed independently by encoder functions (\( f \)), generating compressed latent representations \( Z_{\text{acc}} \) and \( Z_{\text{aud}} \). We explore different fusion approaches, each designed to enhance multi-modal feature extraction in distinct ways.

\begin{enumerate}
    \item \textbf{Joint Variant:}  
    This approach directly concatenates the compressed representations from both modalities, forming a joint feature space that retains their individual characteristics while allowing shared learning:
    \[
    Z_{\text{joint}} = Z_{\text{acc}} + Z_{\text{aud}}
    \]
    The fused representation is then passed into separate decoders to reconstruct the original inputs \( X_{\text{acc'}} \) and \( X_{\text{aud'}} \). This method is particularly effective when both modalities contribute complementary but non-redundant information, allowing the network to learn from each domain while preserving individual modality features. It is well-suited for applications where acceleration and audio signals encode distinct but equally relevant properties of an event.

    \item \textbf{Convolution Variant:}  
    Instead of passing \( Z_{\text{joint}} \) directly to the decoders, this approach applies a convolution operation using a kernel \( K \) to extract deeper cross-modal dependencies:
    \[
    Z_{\text{conv}} = Z_{\text{joint}} \ast K
    \]
    The convolution process enhances interactions between the modalities, capturing shared structural patterns between acceleration and audio features. By leveraging convolutional operations, this method is particularly effective when the modalities exhibit strong interdependencies, such as in impact detection, where motion and sound signals are inherently synchronised. Convolution helps the model recognise localised correlations and refine cross-modal feature extraction, improving robustness in applications where temporal alignment between signals is important.

    \item \textbf{Pooling Variant:}  
    To further refine the fused representation, this variant applies a pooling operation to the convolved features, reducing redundancy and emphasising the most important cross-modal information:
    \[
    Z_{\text{pool}} = \text{Pooling}(Z_{\text{conv}})
    \]
    Pooling ensures that only the most relevant features are retained while reducing computational complexity, making it especially valuable for high-dimensional sensor data. By filtering out less significant features, the model enhances its ability to generalise while improving interpretability. This approach is particularly useful in scenarios where data compression is necessary without losing key information, such as in real-time processing tasks where efficiency is a priority. 
\end{enumerate}

Additionally, two variants were developed based on attention mechanisms drawing inspiration from \citep{Nagrani2021AttentionBF}. \textcolor{black}{Figure \ref{fig:var} provides a visual overview of the architectures used for the different multi-modal variants. The diagram illustrates how accelerometer (\(X_{acc}\)) and audio (\(X_{aud}\)) inputs are encoded into a shared latent representation (\(Z\)), from which various fusion strategies---\textit{Joint}, \textit{Convolution}, \textit{Pooling}, \textit{Attention}, and \textit{Bottleneck Fusion} are derived.} By employing these different fusion mechanisms, the model effectively captures cross-modal interactions between acceleration and audio signals, resulting in a more robust multi-modal learning framework.

\begin{figure*}[ht]
    \centering
    \includegraphics[width=0.98\textwidth]{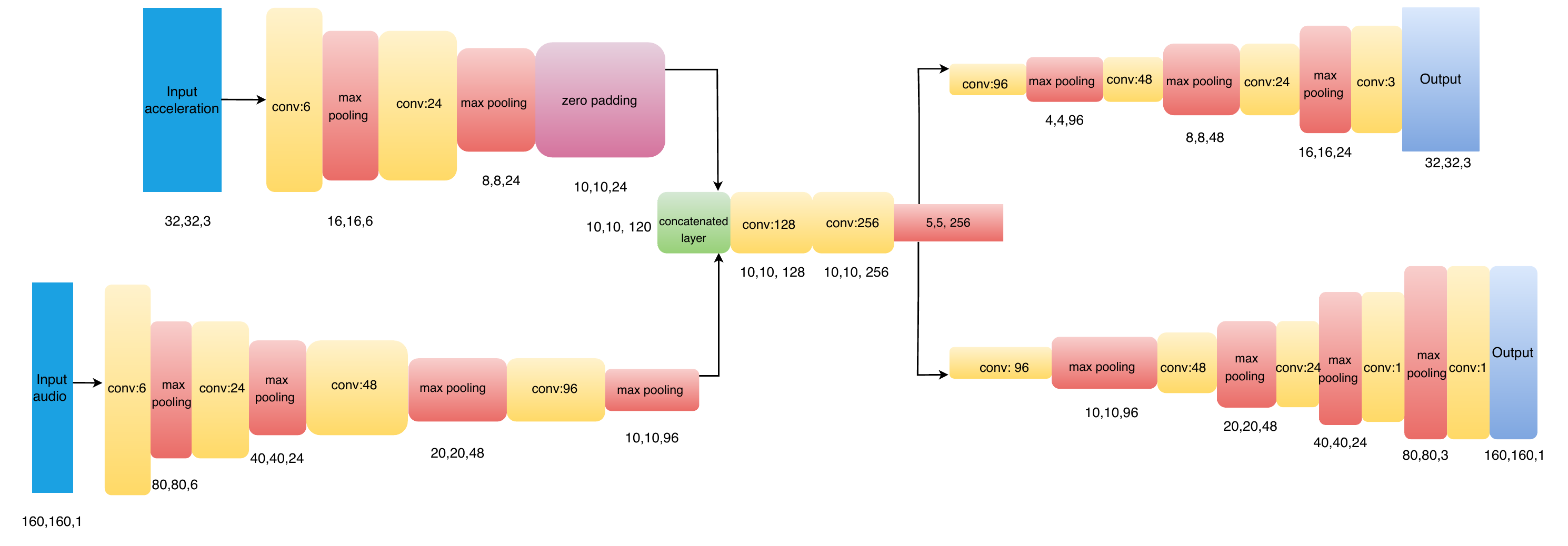}  
    \caption{\textcolor{black}{Architecture of the best-performing pooling-based multi-modal model (MAA3). 
    Separate encoders for acceleration and audio are concatenated into a joint latent space, followed by convolution and pooling operations that enhance cross-modal feature integration.}}
    \label{fig:maa3_architecture}
\end{figure*}

Among the explored fusion strategies, the pooling variant offers a deeper representation of cross-modal feature interactions. Figure \ref{fig:maa3_architecture} presents a deeper view of this architecture, highlighting how modality-specific encoders and subsequent pooling operations enable effective cross-modal feature integration. This model employs separate encoders for acceleration and audio signals, which extract modality-specific representations before merging them into a shared latent space. The fused features are further refined through convolutional and pooling layers, allowing the model to capture complementary information across modalities and enhancing its robustness in small damage detection.

\section{Experimental Evaluation} \label{sec:experiments}
The primary dataset utilised in this study was collected through in-house field testing, with a comprehensive description provided in previous work \citep{sara2024}. The problem was modelled as an anomaly detection task, where autoencoders, a dimensionality reduction technique, were adapted for this purpose. Spectrogram images of acceleration (across all three axes) and audio were input into the autoencoder architecture and its variants, which reconstructed the spectrogram images using the trained model. The reconstruction error was computed, and if it exceeded a predefined threshold, the sample was labelled as an anomaly; otherwise, it was labelled as a background sample.

The model was trained exclusively on samples representing a specific type of damage and subsequently tested on a mixture of background (normal) samples and various other damage samples. The initial dataset comprised approximately 500 instances of damage events, specifically involving vehicle collisions with other objects. Using data augmentation techniques, including rotation and flipping, the dataset was expanded to approximately 3,500 samples for training purposes. The proposed architecture was evaluated on a test set consisting of around 30,000 samples, 20\% of which were damage events. Each sample contained spectrograms of acceleration along the X, Y, and Z axes, as well as audio spectrograms within the 2-3 kHz bandwidth.

\begin{figure*}[htbp]
    \centering
    \includegraphics[width=\textwidth]{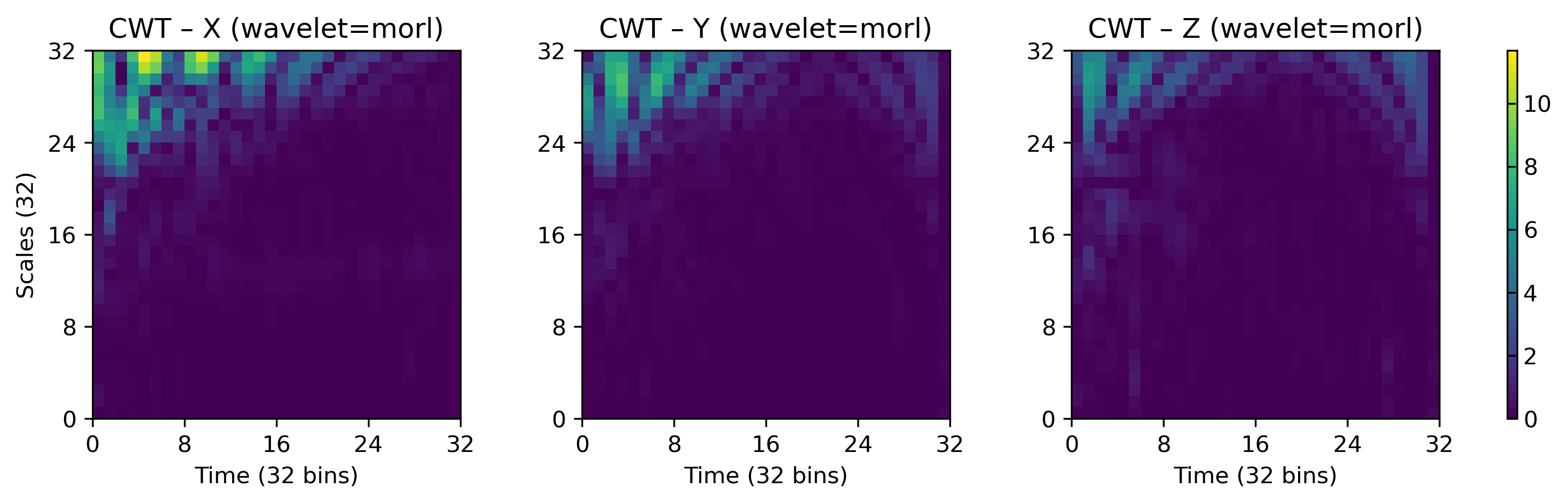}
    \caption{Continuous Wavelet Transform (CWT) spectrograms of acceleration signals along the X, Y, and Z axes during an underbody damage event. The 32×32 time–frequency representations reveal transient energy patterns across scales, indicating structural anomalies.}
    \label{fig:cwt}
\end{figure*}

The implementation was carried out using the Python programming language 3.0 \citep{Python} and TensorFlow library 2.0 \citep{Tensorflow}, with computational resources provided by a Tesla GPU with 32 GB of memory. The signals were transformed into time–frequency representations using the Continuous Wavelet Transform (CWT) \citep{cwt}, which is derived from the Fourier transform. \textcolor{black}{Figure \ref{fig:cwt} illustrates the resulting spectrograms for the X, Y, and Z acceleration axes during an underbody damage event. The colour intensity represents the signal energy across different time and scale bins, where high-energy regions correspond to transient impact responses. This visualisation highlights how wavelet-based analysis reveals structural anomalies that may not be evident in raw time-domain signals.}
 Various optimisers \citep{optimizer}, including Adam, Stochastic Gradient Descent (SGD), and Adadelta, were initially tested on a small subset of samples. Based on the training-validation loss curves, the Adam optimiser was selected, and all results in this work were obtained using it. The model was trained for 200 epochs, with checkpoints saving the model exhibiting the minimum validation loss. The checkpointed model was then loaded for predictions on the test data. A learning rate \citep{learningrate} of 10\textsuperscript{-3} was used for this study. The autoencoder architecture consisted of three layers with 256, 128, and 64 neurons, respectively. Each layer included a convolutional layer, pooling, followed by batch normalisation and an activation function. Model hyperparameters, including fusion depth and network layer configurations, were optimised through a manual grid search. The tuning process focused on minimising training and validation loss, helping to ensure stable convergence and improved generalization performance. Different loss functions, as described in Section \ref{sec:loss}, were evaluated on validation data for a single mono-architecture. 
We used the ROC-AUC curve as the evaluation metric. Since our data is highly imbalanced, it was important to choose an appropriate metric. The ROC-AUC curve plots the True Positive Rate (TPR) against the False Positive Rate (FPR) across different thresholds, allowing us to assess the model's discriminative power. AUC is particularly well-suited for anomaly detection because it handles imbalanced data effectively and is threshold-independent, meaning it does not rely on a specific decision threshold. Instead, it evaluates the model's ability to rank anomalies correctly. While other metrics, such as Balanced Accuracy \citep{balanced}, also address class imbalance, AUC's ranking capability, i.e., ranking data points from \textit{most likely to be anomalous} to \textit{least likely}, is more critical for anomaly detection than exact classification and may support further extensions of this work.

\subsection{Loss Functions} \label{sec:loss}
This section discusses several widely used loss functions in machine learning and image processing, including Mean Squared Error (MSE), Structural Similarity Index (SSIM), Mean Squared Logarithmic Error (MSLE), and Log-Cosh Loss \citep{logcosh}. Each loss function serves a distinct purpose depending on the task, whether regression, image reconstruction, or anomaly detection. Additionally, some models incorporate specialised loss components, such as disentangling losses with regularisation terms (e.g., sparsity) to promote interpretable and efficient latent representations. The following notations are used: \(y_i\) represents the actual value, \(\hat{y}_i\) the predicted value, and \(N\) the number of samples.

\subsubsection{Mean Squared Error (MSE)}
MSE is commonly used in regression problems to measure how far predictions deviate from actual values. It calculates the mean of the squared differences between them:
\[
\text{MSE} = \frac{1}{n} \sum_{i=1}^{n} \left( y_i - \hat{y}_i \right)^2
\]
MSE is straightforward to implement and optimises easily. However, it heavily penalises large errors due to the squared term, which can be problematic if the dataset contains outliers.

\subsubsection{Structural Similarity Index (SSIM)}
SSIM \citep{ssim} is a metric for comparing perceptual similarity between two images. It is especially useful in image reconstruction tasks where maintaining visual quality is important. SSIM evaluates three components: luminance, contrast, and structure:
\[
\text{SSIM}(x, y) = \frac{(2\mu_x \mu_y + C_1)(2\sigma_{xy} + C_2)}{(\mu_x^2 + \mu_y^2 + C_1)(\sigma_x^2 + \sigma_y^2 + C_2)}
\]
Here, \( \mu_x \) and \( \mu_y \) are the means of the images, \( \sigma_x^2 \) and \( \sigma_y^2 \) their variances, \( \sigma_{xy} \) their covariance, and \(C_1\), \(C_2\) constants to prevent division by zero.

Unlike pixel-wise metrics such as MSE, SSIM aligns more closely with human perception, making it suitable for tasks where image quality is critical. However, it is computationally more expensive and less applicable to non-image tasks.

\subsubsection{Mean Squared Logarithmic Error (MSLE)}
MSLE is designed for regression problems where the target variable spans multiple orders of magnitude. It computes the logarithmic difference between the predicted and actual values:
\[
\text{MSLE} = \frac{1}{n} \sum_{i=1}^{n} \left( \log(1 + y_i) - \log(1 + \hat{y}_i) \right)^2
\]
By focusing on proportional differences, MSLE reduces the influence of large absolute errors, making it suitable for tasks where relative accuracy is more important. However, it performs poorly with small-scale errors and requires all target values to be non-negative.

\subsubsection{Log-Cosh Loss}
Log-Cosh is a robust alternative to MSE for regression tasks, as it is less sensitive to outliers \citep{logcosh}. It is defined as:
\[
\text{Log-Cosh} = \sum_{i=1}^{n} \log\left(\cosh\left(\hat{y}_i - y_i\right)\right)
\]
This function smooths the impact of large differences, preventing them from dominating the loss. While effective in handling extreme values, Log-Cosh can be harder to interpret than simpler metrics and may require longer computation times for large datasets.

\subsubsection{Disentangling Loss Components}
In certain models, such as autoencoders, additional loss components encourage the learning of simpler and more interpretable representations. For instance, sparsity regularisation ensures that only a small subset of neurons activate at a time, resulting in compact and efficient representations. 

Sparse autoencoders, for example, include a penalty for excessive activations, which makes the learned features more meaningful and improves generalisation \citep{disent}.

\section{Results and Discussion}\label{sec:results}
This section discusses the results and analysis of the proposed models and loss functions. We evaluated various loss functions on the validation dataset to improve model performance. Initially, MSE was chosen due to its simplicity and effectiveness in regression tasks. However, the performance was suboptimal, prompting exploration of logarithmic loss functions such as Log-Cosh, which handle outliers more robustly and provide smoother gradients. Additionally, SSIM loss was implemented, as it assesses structural similarity rather than point-wise errors, offering a perceptual perspective on image quality.

\begin{figure}[h]
    \centering
    \scalebox{1.1}{ 
        \includegraphics[width=0.45\textwidth, keepaspectratio]{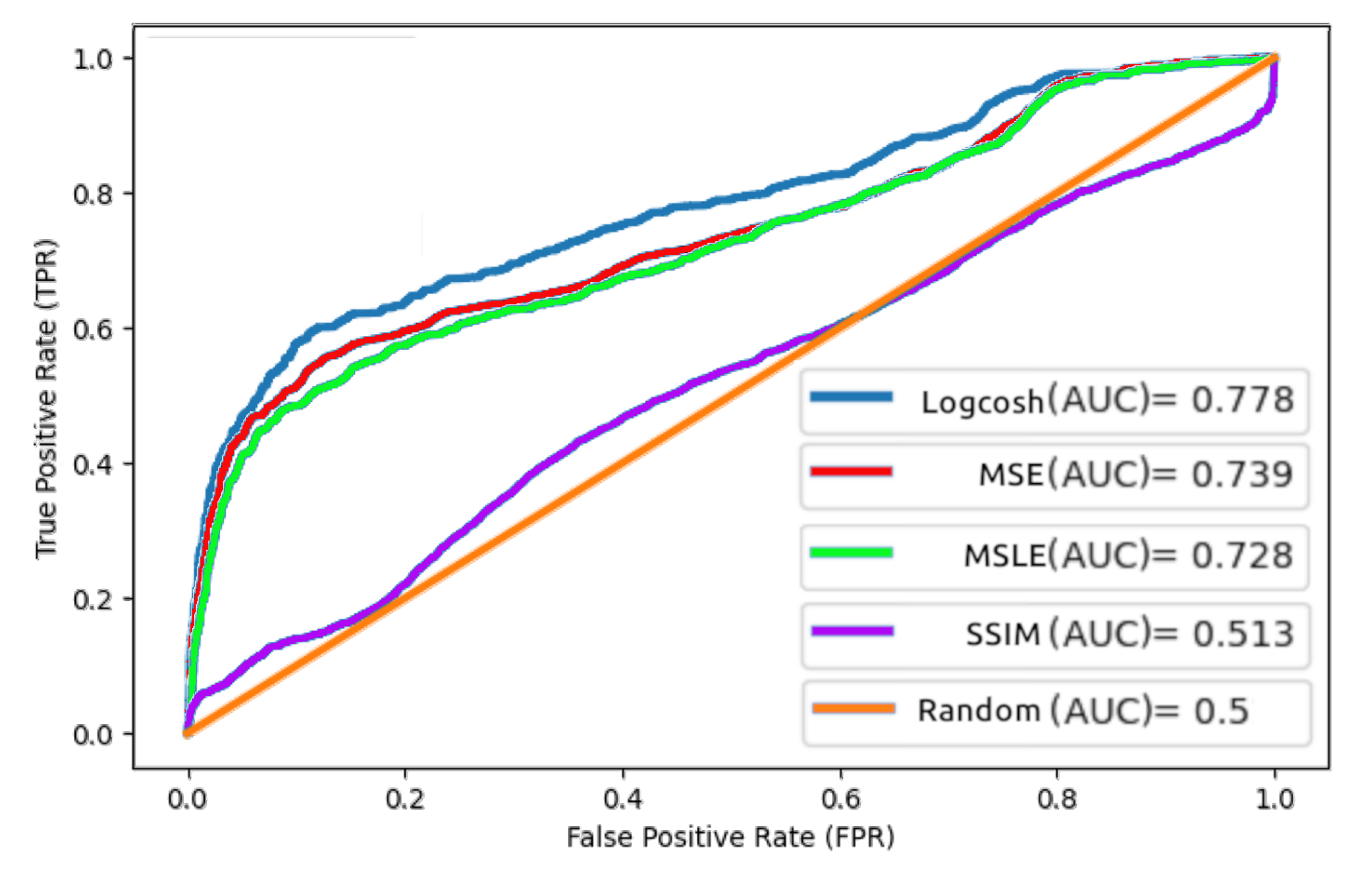}
    }
    \caption{\textbf{ROC-AUC Curves for Different Loss Functions.}  
    The blue, red, green, and purple curves represent Logcosh, MSE, MSLE, and SSIM losses, respectively. Logcosh achieves the highest AUC (0.778), while SSIM performs the worst (AUC = 0.513).}
    \label{fig:roc_auc22}
\end{figure}

Based on the evaluation of different loss functions on the validation dataset, distinct performance variations were observed, as shown in Figure \ref{fig:roc_auc22}. The Log-Cosh loss function achieved the highest AUC of 0.778, indicating its robustness in handling outliers and providing stable gradients, which likely contributed to its superior performance. MSE followed with an AUC of 0.739, demonstrating effectiveness but also sensitivity to outliers, which may have affected results. MSLE achieved an AUC of 0.728, suggesting that while it can be beneficial for targets exhibiting exponential growth, it did not outperform Log-Cosh in this context. Lastly, SSIM loss had the lowest AUC of 0.513. Although SSIM excels at capturing structural similarity, it may not align well with the ROC–AUC metric, which focuses on classification performance. Overall, Log-Cosh emerged as the most effective loss function, balancing outlier handling and maintaining smooth learning dynamics.

We also compared our models in mono-modal and multi-modal configurations using the same test dataset. The mono-modal models for acceleration and audio are denoted by \textit{MAcc} and \textit{MAud}, respectively. The joint, convolutional, and pooling variants are denoted by \textit{MAA1}, \textit{MAA2}, and \textit{MAA3}. Furthermore, following the work of \citep{Nagrani2021AttentionBF}, two attention units were introduced between three layers to form a bottleneck fusion architecture, denoted by \textit{MBotF}. An attention layer applied to the joint representation is denoted by \textit{MAtten}.

\textcolor{black}{From an ethical and privacy perspective, all data collection and processing were conducted in accordance with its internal data-protection policies and GDPR regulations. No raw audio recordings or acoustic features were transmitted or stored externally; all audio processing and feature extraction occurred locally on the sensing device. Only derived IMU features were transmitted to the cloud, and only when a potential damage event was detected. This design minimises privacy risks and ensures responsible, compliant use of sensor data within the operational environment.}

\textcolor{black}{Table \ref{tbl:modelsize} compares model size and inference time across multi-modal architectures. MAA3 offers the best balance, with a size of 12.02 MB and an inference time of 28.8 ms, while the bottleneck fusion model (MBotF) has a similar size but much higher latency, limiting real-time use. Single-modality models are smaller and faster but less accurate. These results suggest that mid-fusion models such as MAA3 are well-suited for efficient and accurate damage detection. Although the architectures were developed and trained on a Tesla V100 GPU, inference times in Table \ref{tbl:modelsize} were measured on a standard CPU setup (8 GB RAM) to reflect typical edge-device performance. Overall, increasing model complexity yields diminishing returns relative to computational cost, highlighting the efficiency advantage of mid-fusion architectures for embedded deployment.}

\begin{table}[h]  
\centering
\caption{\textbf{Comparison of Multi-Modal Architectures Based on Model Size and Inference Time.}  
Model inference times are averaged per sample. \textbf{MAA3} achieves the best balance between performance and latency, while the bottleneck fusion variant shows significant latency despite a similar model size.}
\label{tbl:modelsize}
\begin{tabular}{lcc}  
\toprule
\textbf{Models} & \textbf{Model Size (MB)} & \textbf{Inference Time (ms/sample)} \\ 
\midrule
MAcc  & 1.29  & 6.331 \\
MAud  & 8.50  & 21.304 \\
MAA1  & 6.87  & 22.584 \\
MAA2  & 6.11  & 23.513 \\
MAA3  & \textbf{12.02} & \textbf{28.755} \\
MBotF & 12.37 & 322.727 \\
\bottomrule
\end{tabular}
\end{table}

\begin{table}[h]
\centering
\caption{\textbf{Comparison of Multi-Modal Architectures Based on ROC-AUC Scores.}  
The highest ROC-AUC score is achieved by \textbf{MAA3} (0.92) using audio modality, while MAcc and MAud perform best in their respective single-modal settings. Multi-modal architectures generally show improvements over single-modality approaches.}
\label{tbl2}
\resizebox{\columnwidth}{!}{
\begin{tabular}{cccc}
\toprule
\textbf{Models} & \textbf{ROC-AUC (Acc.)} & \textbf{ROC-AUC (Au.)} & \textbf{ROC-AUC (Best)} \\
\midrule
MAcc & \textbf{0.80} & N/A & 0.80 \\
MAud & N/A & 0.84 & 0.84 \\
MAA1 & 0.46 & 0.74 & 0.74 \\
MAA2 & 0.64 & 0.82 & 0.82 \\
MAA3 & 0.67 & \textbf{0.92} & \textbf{0.92} \\
MBotF & 0.55 & 0.62 & 0.62 \\
MAtten & 0.53 & 0.62 & 0.62 \\
\bottomrule
\end{tabular}
}
\end{table}

\subsection{Analysis of Results}
This section evaluates the performance of the proposed models and compares their effectiveness across different sensing modalities. 
The results presented in Table \ref{tbl2} highlight the performance of different models on the acceleration (Acc.) and audio (Aud.) modalities. The models were compared on a test dataset using ROC–AUC metric. A value close to 1 indicates high discriminative power between damage and non-damage events, whereas a score near 0.5 indicates no discriminative power at all \citep{roc}. This analysis considers both the architectural impacts and the challenges posed by the data.

\subsubsection{Mono-Modal Models}
\begin{itemize}
    \item \textbf{MAcc:} Using only acceleration data, this model achieved an ROC–AUC score of 0.80, indicating efficient feature extraction from acceleration data alone, which proved effective in certain scenarios.
    \item \textbf{MAud:} The audio-only model performed slightly better, with an ROC–AUC of 0.84, suggesting that audio data may provide more discriminative features for the task, likely due to its richer and more varied informational content.
\end{itemize}

\textcolor{black}{These results from Table~\ref{tbl2} indicate that while both single-modality models perform well, the audio-based approach (MAud) captures richer discriminative cues than the acceleration-based model (MAcc).}

\begin{figure*}[t]
    \centering
    \begin{subfigure}[t]{0.48\textwidth}
        \centering
        \includegraphics[width=\linewidth]{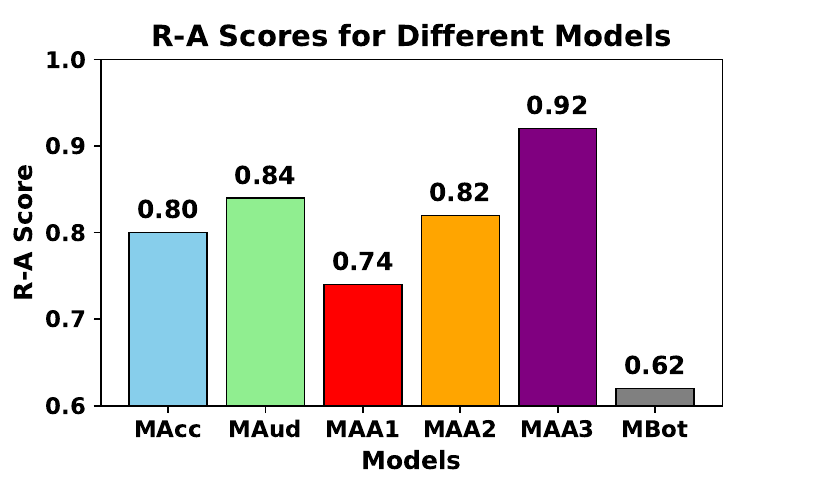}
        \caption{Comparison of ROC–AUC scores for different models. 
        Model MAA3 achieves the highest score (0.92), while MBot performs the lowest (0.62). 
        The results indicate that incorporating both acceleration and audio modalities improves performance.}
        \label{FIG:roc_auc1}
    \end{subfigure}
    \hfill
    \begin{subfigure}[t]{0.48\textwidth}
        \centering
        \includegraphics[width=\linewidth]{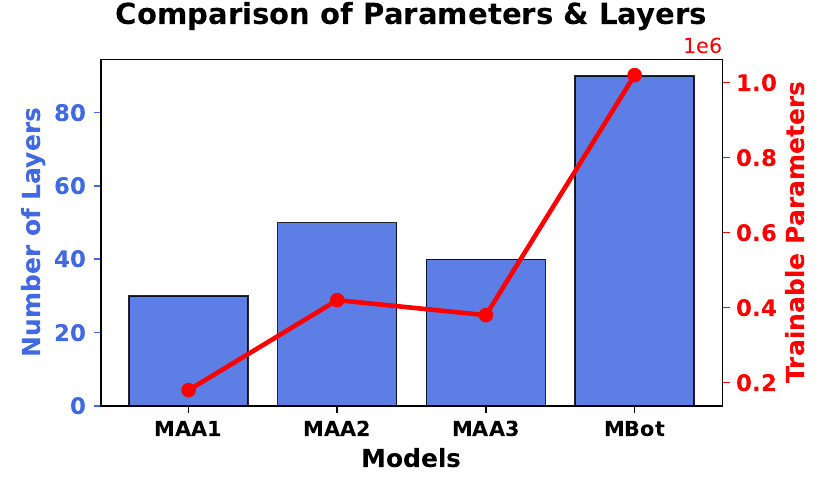}
        \caption{Comparison of the number of trainable parameters and layers across different multi-modal architectures. 
        MBot has the highest number of parameters and layers, while MAA1 has the least. 
        This suggests that more complex models require significantly higher computational resources.}
        \label{FIG:multi_modal}
    \end{subfigure}
    \caption{\textbf{Performance and model complexity comparison across architectures.}
    (a) ROC–AUC scores show the effect of modality fusion on detection accuracy. 
    (b) The number of parameters and layers highlights computational trade-offs among architectures.}
    \label{FIG:roc_auc_vs_params}
\end{figure*}

\subsubsection{Multi-Modal Models}
\textcolor{black}{Table~\ref{tbl2} also compares the performance of the multi-modal variants, which explore different strategies for fusing acceleration and audio representations.}

\begin{itemize}
    \item \textbf{MAA1 (Joint Variant):} This model exhibited lower performance, with an ROC–AUC score of 0.46 for acceleration and 0.74 for audio. The direct concatenation of features may not capture complex interdependencies effectively, resulting in poorer performance.
    \item \textbf{MAA2 (Convolutional Variant):} This variant showed improved results, achieving scores of 0.64 for acceleration and 0.82 for audio. The convolutional layers likely enhanced the capture of intricate cross-modal patterns, though performance still lagged behind the mono-modal audio model.
    \item \textbf{MAA3 (Pooling Variant):} This variant achieved the best performance among the multi-modal models, with scores of 0.67 for acceleration and a notable 0.92 for audio. Pooling layers effectively reduced noise and emphasised the most relevant features, helping to distinguish critical patterns within the data.
    \item \textbf{MAtten (Attention Variant):} This model achieved moderate performance with scores of 0.53 for acceleration and 0.62 for audio. Although attention mechanisms are designed to prioritise the most informative features, the model did not substantially outperform others, possibly due to suboptimal alignment or limited training to fully leverage attention.
    \item \textbf{MBotF (Bottleneck Fusion Variant):} This model scored 0.55 for acceleration and 0.62 for audio, indicating that the bottleneck fusion approach did not yield the expected benefits. This may have resulted from over-compression or the loss of crucial information during feature combination.
\end{itemize}

\begin{table}[h]
\centering
\caption{\textbf{Misclassification Summary by Event Name.}  
Number of false positives (FP) for each event category used in the test set. Labels with high FP may indicate ambiguous or challenging damage types.}
\label{tbl:misclassification_summary}
\scriptsize
\begin{tabular}{cc}
\toprule
\textbf{Event Name} & \textbf{False Positives (FP)}  \\
\midrule
None & 748  \\
Pothole & 380 \\
Speed Bump & 65 \\
General Bump & 15  \\
Curb Climb & 8  \\
Vehicle hits Front Right Bumper & 6  \\
Vehicle hits Front Left Bumper & 8 \\
Vehicle hits Back Right Bumper & 6  \\
Vehicle hits Back Left Bumper & 6  \\
Vehicle hits Back Bumper & 6  \\
Vehicle hits Front Bumper & 6  \\
ABT-Bottom-Out & 18  \\
Object Impact Front Left Bumper & 4  \\
Object Impact Front Bumper & 6  \\
Object Impact Back Bumper & 6  \\
Roof Slap Front Left Outside & 12 \\
Door Close Trunk & 23 \\
\bottomrule
\end{tabular}
\end{table}

\subsubsection{Critical Evaluation}
The \textit{MAA3 (Pooling Variant)} model outperformed all others overall, particularly in the audio modality, with an ROC–AUC score of 0.92. This suggests that pooling effectively reduced noise and enhanced multi-modal feature integration. The mono-modal audio model (\textit{MAud}) performed nearly as well as \textit{MAA3}, indicating that although multi-modal integration can be beneficial, it remains essential to address the challenge of similar representations between damage and non-damage samples. The poor performance of \textit{MAA1} highlights the limitations of simple fusion strategies, where concatenation alone fails to capture complex inter-modal relationships.  

The underperformance of \textit{MBotF} illustrates that more complex architectures, such as those employing attention mechanisms, do not necessarily yield superior results and may introduce additional difficulties in model alignment and representation learning. Our analysis revealed that audio features benefited more from multi-modal fusion than acceleration data, suggesting modality-specific differences in how cross-modal information can be exploited effectively. These findings will guide future work towards developing more robust fusion architectures.  

Figure \ref{FIG:roc_auc1} shows the best-performing model based on AUC scores. Using the maximum-selection strategy, the higher ROC–AUC of either acceleration or audio was taken as the representative score. Figure \ref{FIG:multi_modal} provides an overview of the number of trainable parameters and layers across the different multi-modal architectures. Each operation, including convolution, batch normalisation, and max-pooling, is implemented as a separate layer. Due to the autoencoder-based design, redundancy naturally occurs in the decoder during input reconstruction. Overall, the diagram illustrates that the pooling-based architecture achieves an effective balance between complexity and efficiency, which explains its superior performance compared to deeper or more parameter-heavy models.

Table \ref{tbl:misclassification_summary} presents the false-positive (FP) counts for various event categories in the test set using MAA3. False positives occur when the model incorrectly classifies benign or normal events as damage. Analysing which events yield frequent FPs helps identify model limitations. Notably, events such as \textit{Door Close}, \textit{Pothole}, and \textit{Speed Bump} show higher FP rates. This is likely because these events generate signal patterns, sudden accelerations or vibrations that closely resemble those of actual damage. For instance, acceleration spikes produced when a door closes can mimic impact signatures, causing the model to confuse them with damage events. Such signal similarities highlight the challenge of distinguishing subtle nuances between non-damage but abrupt vehicle motions and true damage occurrences. These insights suggest that future research should focus on improving the model’s ability to discriminate between these confounding events.

\begin{figure*}[t]
    \centering
    \begin{subfigure}[b]{0.49\textwidth}
        \centering
        \includegraphics[width=\linewidth]{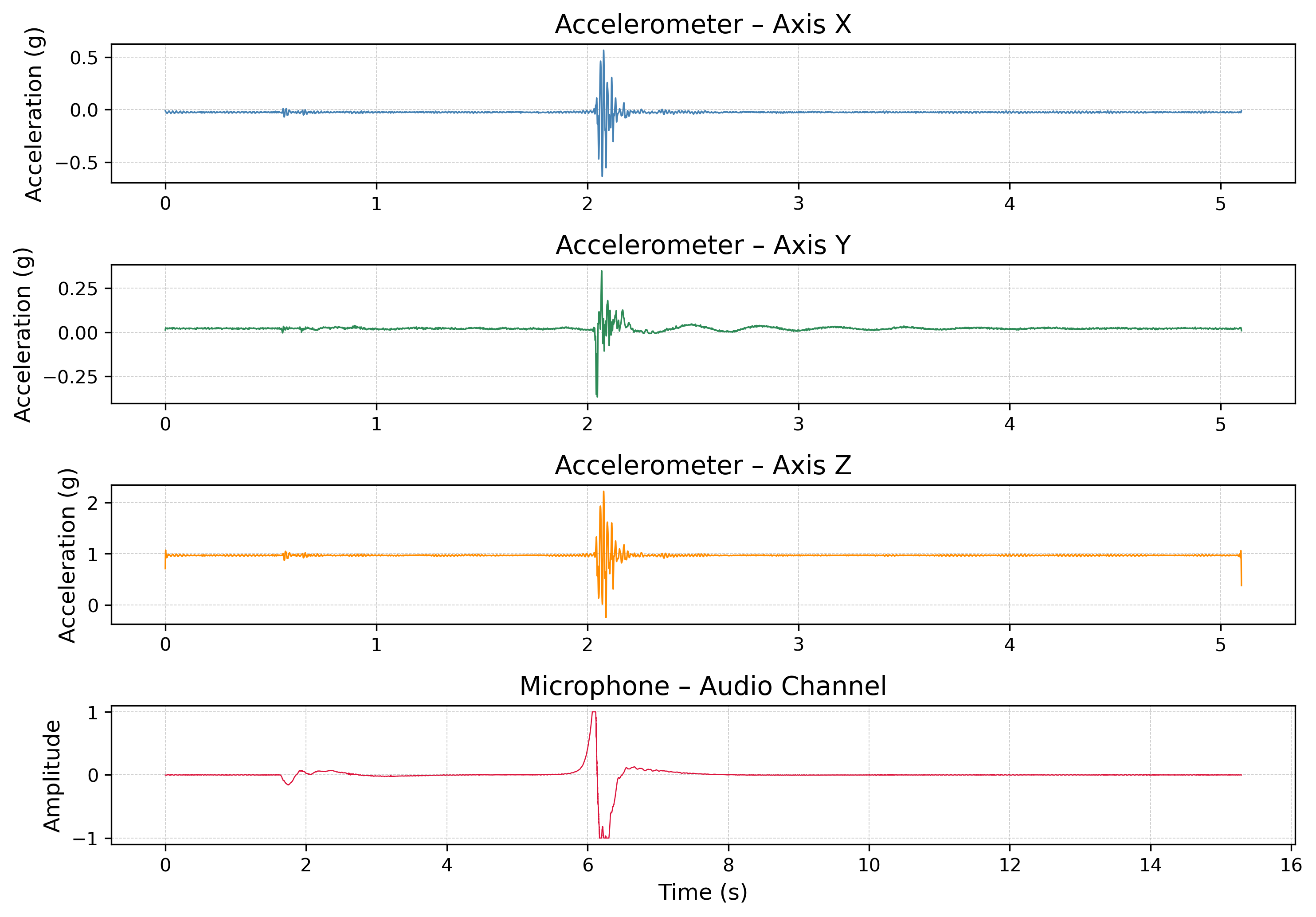}
        \caption{\textbf{Door Closing Event.}
        The accelerometer (X, Y, Z) and audio channels show a short, high-amplitude transient caused by door impact. 
        This sharp vibration pattern closely resembles that of a genuine collision, leading to false-positive detections.}
        \label{fig:door_close_acc_audio}
    \end{subfigure}
    \hfill
    \begin{subfigure}[b]{0.49\textwidth}
        \centering
        \includegraphics[width=\linewidth]{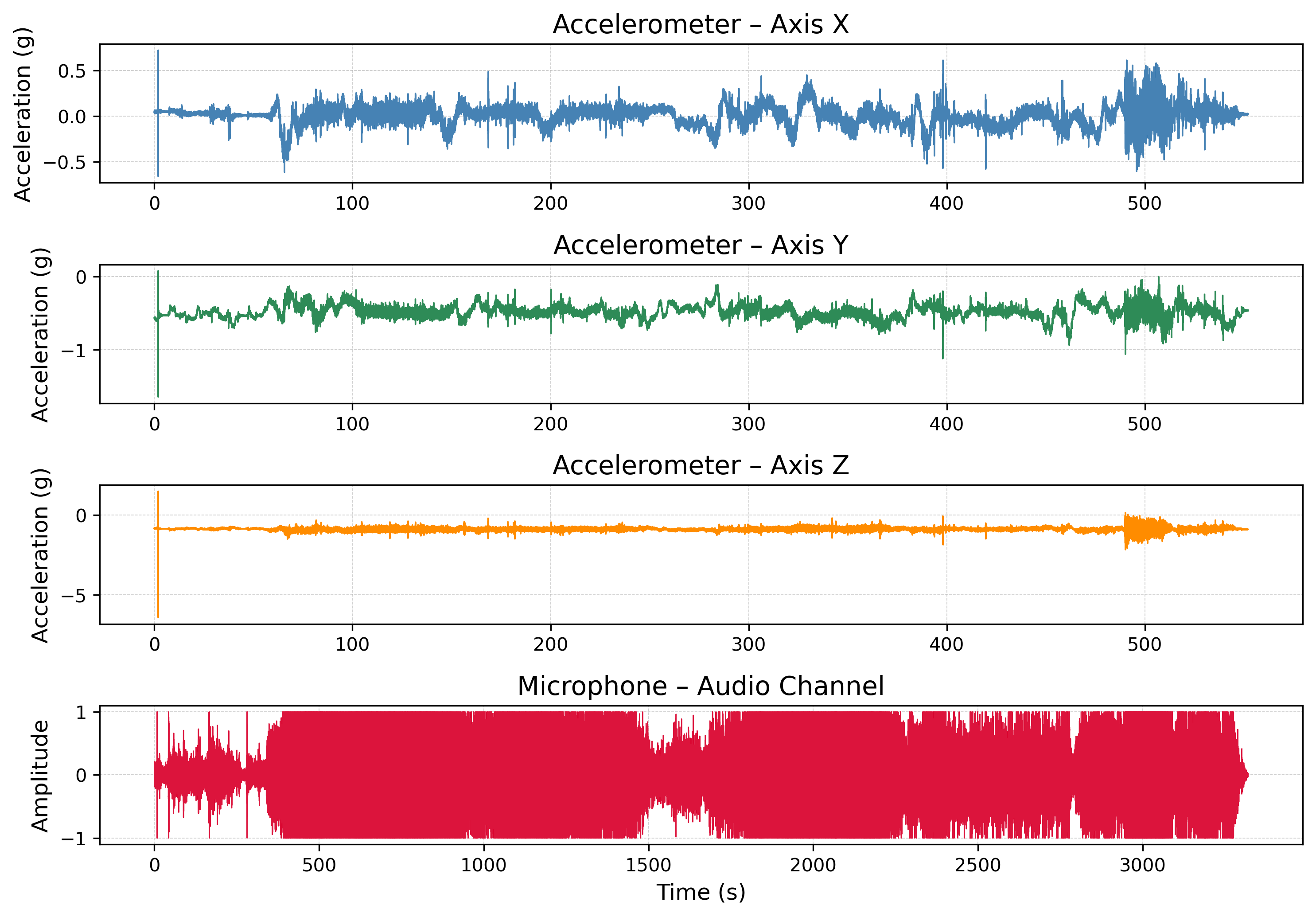}
        \caption{\textbf{Random Driving Event.}
        Typical non-damage vibrations recorded during regular driving. 
        The motion and acoustic activity are continuous and broadband, without distinct impulsive transients. 
        The model correctly identifies this sequence as non-damage.}
        \label{fig:random_drive_acc_audio}
    \end{subfigure}
    \caption{\textbf{Qualitative analysis of error and non-error cases.}
    Comparison of (a) a misclassified event and (b) a correctly classified non-damage event. 
    The door-closing sequence exhibits brief impact-like transients that the model confuses with damage,
    while the random driving case shows smoother, distributed motion signatures. 
    These examples visually explain the high false-positive counts reported in Table~\ref{tbl:misclassification_summary}.}
    \label{fig:qualitative_fp_cases}
\end{figure*}

\textcolor{black}{{Figures \ref{fig:door_close_acc_audio} and \ref{fig:random_drive_acc_audio} illustrate two representative examples of such misclassified or confounding cases.
Figure \ref{fig:door_close_acc_audio} shows the raw accelerometer and audio signals recorded during a door-closing event. A brief but high-amplitude transient is visible across all accelerometer axes, accompanied by a sharp acoustic spike, features that strongly resemble those of a genuine impact.
By contrast, Figure \ref{fig:random_drive_acc_audio} depicts an extended random driving sequence with no damage occurrence. Here, the vibration signatures appear more distributed and broadband, yet brief motion bursts occasionally reach amplitudes similar to those seen in true impact events.}
These qualitative examples visually confirm the patterns behind the false-positive counts reported in Table \ref{tbl:misclassification_summary}. The resemblance between short, high-energy transients from everyday vehicle interactions and genuine impact signals remains a major source of confusion for the model. Incorporating temporal context or multi-event reasoning, such as verifying consistency across multiple time windows or fusing inertial and acoustic cues more tightly, could mitigate these errors and further improve the robustness of future versions of MAA3.}

\subsection{Comparison with CVAE and CLARM}
To broaden the analysis, we evaluated recently developed architectures. Convolutional layers were enhanced by integrating residual blocks, and more sophisticated models such as Conditional Variational Autoencoders (CVAE) \citep{cvae} and a hybrid model combining CVAE with Long Short-Term Memory (LSTM) networks, inspired by \citep{Rautela2024}, were investigated. The CVAE model comprised a latent space of two units, whereas the hybrid model employed a CVAE framework augmented with an LSTM containing 64 neurons in its latent space. Table \ref{tbl9} summarises the performance of these advanced multi-modal architectures, illustrating the relative effectiveness of each configuration.

\begin{table}[h]  
\scriptsize
\centering
\caption{\textbf{Comparative Results of Advanced Multi-Modal Architectures.}  
The highest ROC-AUC score is achieved by MAA3 (0.92) using the audio modality, significantly outperforming other models. CVAE and Residual Blocks show moderate performance, but their best scores remain lower than MAA3.}
\label{tbl9}
\begin{tabular}{cccc}  
\toprule
\textbf{Models} & \textbf{ROC-AUC (Acc.) | (Au.)} & \textbf{ROC-AUC (Best)} \\ 
\midrule
MAA3 & 0.67 | \textbf{0.92} & \textbf{0.92} \\
CVAE & 0.51 | 0.61 & 0.61 \\
CVAE(L) & 0.50 | 0.52 & 0.52 \\
Residual Blocks & 0.56 | 0.62 & 0.62 \\
\bottomrule
\end{tabular}
\end{table}

Table \ref{tbl9} further confirms the superior capability of MAA3 in processing audio data, achieving the highest accuracy and AUC metrics. The results for CVAE and its hybrid variant indicate that although incorporating LSTM layers enhances the modelling of temporal dependencies, it does not substantially improve performance compared with the base CVAE. The model with residual blocks achieved moderate gains, suggesting that while these blocks stabilise training for deeper networks, they do not markedly alter overall performance outcomes.

\subsection{Generalisability Evaluation}
To assess cross-domain generalisability, we evaluated our pretrained MAA3 model on an open-source multi-modal anomaly detection dataset for autonomous mobile robots \citep{hao2025unsupervised}. This dataset comprises approximately 4,500 training samples and 2,000 test samples, with the latter including around 800 abnormal events such as collisions and internal malfunctions. Without any retraining or fine-tuning, our model achieved an ROC–AUC of 0.90. After retraining on the 4,500 normal training samples for 30 epochs, contrary to the original training setup and without further hyperparameter optimisation, the performance improved to 0.934 ROC–AUC. This demonstrates that with further fine-tuning, the proposed fusion architecture could approach or even surpass the reported state-of-the-art ROC–AUC of 0.97 on this dataset, highlighting its strong cross-domain adaptability beyond vehicle damage detection.

\section{Conclusion and Future Work} \label{sec:conclusions}
This work presented a sensor-based, data-driven framework for detecting minor vehicle damage in real time, addressing a key limitation of camera-based systems that struggle with underbody or in-motion damage detection. Such capability is essential for decentralised inspection scenarios, such as car rental or shared-mobility applications, where vehicles cannot be manually checked after each use. The proposed system employs a windshield-mounted device equipped with IMU and microphone sensors to capture impact-related signals. Using autoencoder-based anomaly detection, the method reduces the need for extensive labelled datasets and enables unsupervised learning of discriminative representations. Although autoencoders are traditionally used for dimensionality reduction, they can be reformulated to learn normal behaviour and identify anomalies through reconstruction error, allowing reliable detection of subtle damage events.  

A key contribution of this research lies in the design and evaluation of cross-modal fusion architectures that integrate IMU and audio data. The proposed multi-modal approach consistently outperformed mono-modal baselines, demonstrating the value of complementary sensing in capturing different physical aspects of damage events. Furthermore, the comparative analysis of architectural variants showed that increased model depth or parameter count does not necessarily yield better results; for example, adding LSTM layers to the CVAE did not significantly improve results compared with the base configuration. Simple pooling-based fusion strategies provided an effective balance between accuracy and efficiency, making them well-suited for real-time automotive applications.

Future work will focus on refining hyperparameter optimisation to enhance both advanced (e.g., CVAE-based) and lightweight convolutional variants. The current training dataset primarily consists of vehicle–object collision scenarios, but future datasets will incorporate a broader range of damage types and environmental conditions. A staged expansion of training data, introducing new damage categories incrementally, may provide more stable learning dynamics and reveal how specific event classes influence generalisation.  

\textcolor{black}{Although our experiments encompassed both stationary and moving vehicles across varied weather and road conditions, a systematic robustness evaluation under extreme environments remains pending. Differences in terrain, vehicle type, or acoustic context can alter sensor responses and affect detection reliability. Future work will therefore include targeted cross-environment assessments (e.g., snow, heavy rain, and uneven surfaces) and domain-adaptation techniques to ensure stable performance across diverse vehicle fleets.}

Additional efforts will be directed towards improving model interpretability, lowering false positive rates, and optimising inference for edge devices with limited computational resources. Deployment considerations, such as power consumption, latency, and user-interface integration for fleet management, will also form part of future research. Collectively, these directions aim to translate the proposed Small Damage Detection (SDD) framework from controlled experimental settings to scalable, real-world automotive applications.

\textcolor{black}{
\section*{CRediT authorship contribution statement}
Sara Khan: Conceptualisation, Methodology, Experiments, Investigation, Data curation, Visualisation, Writing – Original Draft, Revision.\\
Mehmed Yüksel: Visualisation, Figure preparation, Review.\\
Frank Kirchner: Review and feedback.}

\textcolor{black}{
\section*{Declaration of competing interest}
The authors have no conflicts of interest to declare.
}

\bibliographystyle{model5-names}   
\bibliography{cas-dc-template}

\end{document}